\documentclass[10pt,twocolumn,letterpaper]{article}

\usepackage{cvpr}      % To produce the REVIEW version

\usepackage{color, xcolor} 
\usepackage{colortbl}
\usepackage{graphicx}
\usepackage{adjustbox}
\usepackage{makecell}
\usepackage{multirow}
\usepackage{graphicx}    % 用于插入图片
\usepackage{amsmath}     % 用于数学公式和符号
\usepackage{amssymb}     % 用于数学符号如\mathbb{R}
\usepackage{algorithm}   % 用于算法伪代码环境
\usepackage{algpseudocode}
\usepackage{amsfonts} 
\definecolor{cvprblue}{rgb}{0.21,0.49,0.74}
\usepackage[pagebackref,breaklinks,colorlinks,allcolors=cvprblue]{hyperref}

\def\paperID{} % *** Enter the Paper ID here
\def\confName{CVPR} 
\def\confYear{2026}

\title{SR3R: Rethinking Super-Resolution 3D Reconstruction With Feed-Forward Gaussian Splatting}
\author{Xiang Feng\textsuperscript{\rm 1 \rm 2}\footnotemark[2] \footnotemark[1] \quad  Xiangbo Wang\textsuperscript{\rm 1}\footnotemark[1] \quad  Tieshi Zhong \textsuperscript{\rm 1}   \quad Chengkai Wang\textsuperscript{\rm 1}\quad Yiting Zhao \textsuperscript{\rm 1}   \quad Tianxiang Xu \textsuperscript{\rm 4}   \quad \\ Zhenzhong Kuang \textsuperscript{\rm 1} \footnotemark[3]  \quad Feiwei Qin \textsuperscript{\rm 1}   \quad Xuefei Yin \textsuperscript{\rm 3}   \quad Yanming Zhu \textsuperscript{\rm 3} \footnotemark[3]  \quad\\
{\fontsize{10}{15}\selectfont \textsuperscript{\rm 1}Hangzhou Dianzi University} \quad 
{\fontsize{10}{15}\selectfont \textsuperscript{\rm 2}ShanghaiTech University} \quad
{\fontsize{10}{15}\selectfont \textsuperscript{\rm 3}Griffith University} \quad
{\fontsize{10}{15}\selectfont \textsuperscript{\rm 4}Peking University}
\\
{\small \href{https://xiangfeng66.github.io/SR3R/}{https://xiangfeng66.github.io/SR3R/}}
}

\begin{document}
% \maketitle

\twocolumn[{
\renewcommand\twocolumn[1][]{#1}
\maketitle
\begin{center}
    \centering
    \vspace{-15pt}
    \includegraphics[width=\linewidth]{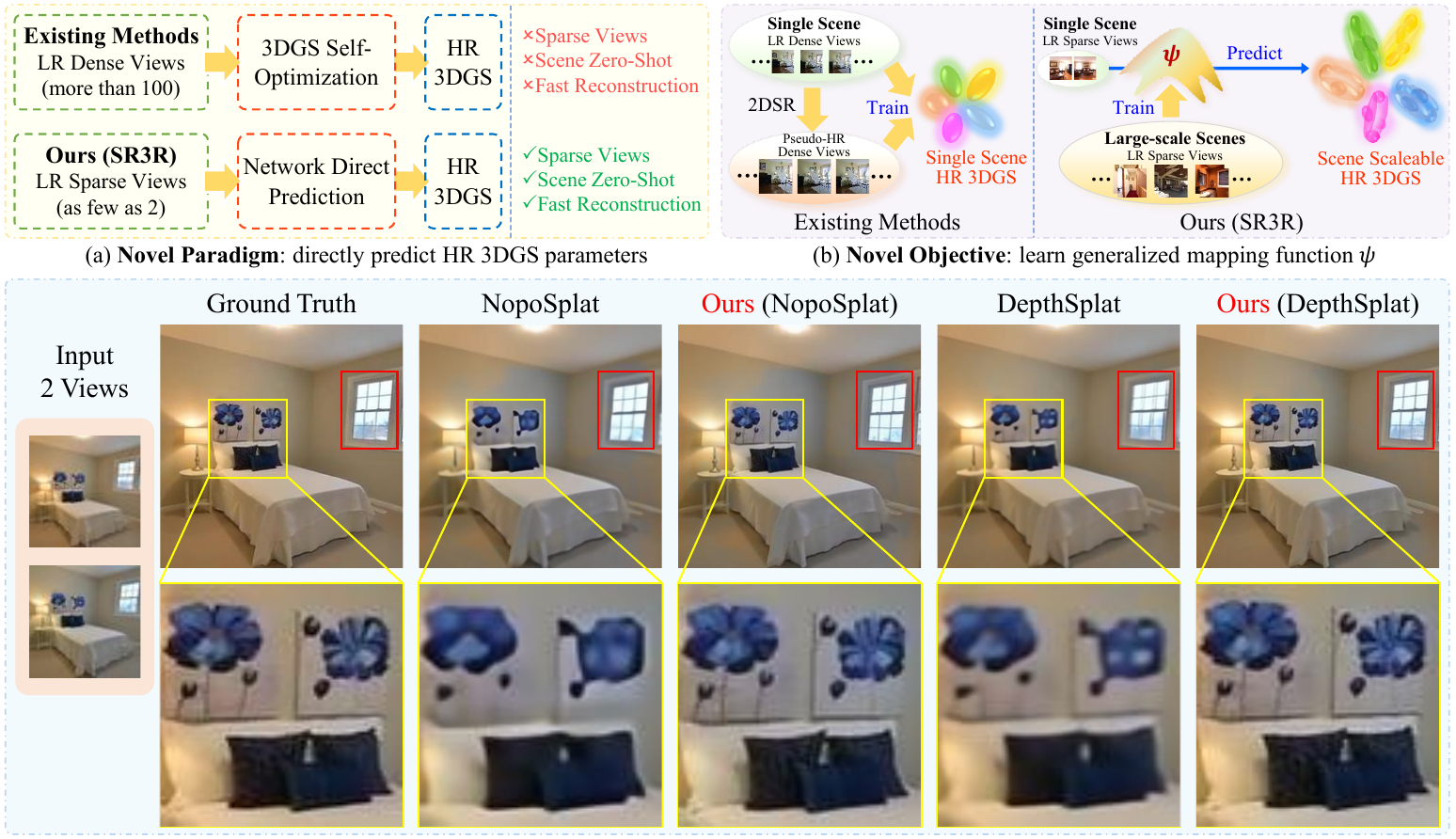}
    \vspace{-15pt}
    \captionof{figure}{We reformulate 3DGS-based 3DSR as a feed-forward mapping problem from sparse LR views to HR 3DGS representation. (a) Unlike existing methods that rely on dense multi-view inputs and per-scene 3DGS self-optimization, our method directly predicts HR 3DGS by a learned network from as few as two LR views. (b) This reformulation fundamentally changes how 3DSR acquires high-frequency knowledge. Instead of inheriting the limited priors embedded in 2DSR models, our SR3R learns a generalized cross-scene mapping function from large-scale multi-scene data, enabling the network to autonomously acquire the 3D-specific high-frequency structures required for accurate HR 3DGS reconstruction. The bottom row illustrates that our SR3R produces significantly sharp and faithful reconstructions.}
    \label{fig:teaser}
\end{center}
}]

\renewcommand{\thefootnote}{\fnsymbol{footnote}}
\footnotetext[1]{Equal contribution}
\footnotetext[2]{Project leader}
\footnotetext[3]{
Corresponding author}

\begin{abstract}

3D super-resolution (3DSR) aims to reconstruct high-resolution (HR) 3D scenes from low-resolution (LR) multi-view images. Existing methods rely on dense LR inputs and per-scene optimization, which restricts the high-frequency priors for constructing HR 3D Gaussian Splatting (3DGS) to those inherited from pretrained 2D super-resolution (2DSR) models. This severely limits reconstruction fidelity, cross-scene generalization, and real-time usability. We propose to reformulate 3DSR as a direct feed-forward mapping from sparse LR views to HR 3DGS representations, enabling the model to autonomously learn 3D-specific high-frequency geometry and appearance from large-scale, multi-scene data. This fundamentally changes how 3DSR acquires high-frequency knowledge and enables robust generalization to unseen scenes. Specifically, we introduce \textbf{SR3R}, a feed-forward framework that directly predicts HR 3DGS representations from sparse LR views via the learned mapping network. To further enhance reconstruction fidelity, we introduce Gaussian offset learning and feature refinement, which stabilize reconstruction and sharpen high-frequency details. SR3R is plug-and-play and can be paired with any feed-forward 3DGS reconstruction backbone: the backbone provides an LR 3DGS scaffold, and SR3R upscales it to an HR 3DGS. Extensive experiments across three 3D benchmarks demonstrate that SR3R surpasses state-of-the-art (SOTA) 3DSR methods and achieves strong zero-shot generalization, even outperforming SOTA per-scene optimization methods on unseen scenes.

\end{abstract}    
\section{Introduction}
\label{sec:intro}
%%3D super-resolution (3DSR) aims to reconstruct high-resolution (HR) 3D representations from low-resolution (LR) multi-view observations. This task is increasingly important in applications such as virtual and augmented reality, autonomous systems, and immersive content creation, where acquiring HR data is often infeasible due to sensor limitations, constrained acquisition, or storage overhead \cite{super-nerf,wan2025s2gaussian}. Among various 3D scene representations, 3D Gaussian Splatting (3DGS) \cite{3Dgaussians} has recently established itself as a leading approach, owing to its real-time rendering capability, continuous representation, and compact storage. These advantages make 3DGS a compelling target for 3DSR, where the objective is to reconstruct HR 3DGS from sparse LR images to enable high-fidelity rendering.% and robust downstream processing.
3D super-resolution (3DSR) aims to reconstruct high-resolution (HR) 3D representations from low-resolution (LR) multi-view observations. This task has become increasingly critical because state-of-the-art 3D Gaussian Splatting (3DGS)–based reconstruction methods \cite{3Dgaussians} typically require dense and high-resolution input views to recover fine geometric and appearance details. However, in real-world scenarios, obtaining such high-quality observations is often infeasible due to sensor resolution limits, constrained capture conditions, and storage or bandwidth restrictions \cite{super-nerf,wan2025s2gaussian}. These practical limitations motivate the development of 3DSR methods capable of lifting sparse and LR inputs to high-fidelity 3D representations.

Current 3DSR methods \cite{feng2024srgs,gaussiansr,sequence,SuperGaussian} typically employ pretrained 2D image or video super-resolution (2DSR) models to generate pseudo-HR images from dense multi-view LR inputs, which are then used as supervision for per-scene optimization of HR 3DGS. Although this strategy injects high-frequency cues into the HR 3DGS reconstruction, it suffers from several fundamental limitations. First, per-scene optimization isolates each scene as an independent problem and restricts the source of high-frequency knowledge to the priors embedded in pretrained 2DSR models. This prevents leveraging large-scale cross-scene data to learn 3D-specific SR priors and to train a generalized 3DSR model, thereby inherently limiting reconstruction fidelity, cross-scene generalization, and real-time usage. Second, reliance on 2DSR-generated pseudo-HR labels inherently caps the achievable reconstruction fidelity. Third, dense multi-view synthesis and iterative optimization introduce substantial computational and data overhead.
%%First, generating reliable pseudo-HR labels requires dense multi-view coverage, resulting in substantial data and computational overhead. Second, the optimization of 3DGS must be performed separately for each scene, making it slow, memory-intensive, and impractical in real-time or large-scale settings. Third, 2D pseudo-labels lack view consistency and fail to capture scene-level geometric coherence, often leading to texture artifacts and geometric distortions during optimization. These compounded issues not only constrain reconstruction fidelity to the quality of 2D supervision but also isolate each scene as an independent optimization problem, preventing the use of cross-scene data to train a generalized 3DSR model.
%prevent cross-scene learning, precluding the possibility of training a generalized 3DSR model.
%preserve the underlying 3D geometry
%First, it requires dense multi-view coverage to produce reliable pseudo-HR labels, incurring significant data and computation costs. Second, it relies on per-scene optimization of 3DGS, which is slow, memory-intensive, and incompatible with real-time or large-scale settings. Third, 2D pseudo-labels lack view consistency and do not capture scene-level geometric coherence, often leading to texture inconsistencies or geometric distortions during 3DGS optimization. 

To address these limitations, we propose \textbf{SR3R}, a feed-forward 3DSR framework that directly predicts HR 3DGS from sparse LR views via a learned mapping network. The key idea behind SR3R is to reformulate 3DSR as a direct mapping from LR views to HR 3DGS representation, enabling the model to autonomously learn high-frequency geometric and texture details from large-scale, multi-scene data. This reformulation replaces the conventional 2DSR prior injection with data-driven 3DSR prior learning, marking a fundamental paradigm shift from per-scene HR 3DGS optimization to generalized HR 3DGS prediction (Fig. \ref{fig:teaser}). Concretely, SR3R first employs any feed-forward 3DGS reconstruction model to estimate an LR 3DGS scaffold from sparse LR views, and then upscales it to HR 3DGS via the learned mapping network. The framework is fully plug-and-play and compatible with existing feed-forward 3DGS pipelines. To further enhance reconstruction fidelity, we introduce Gaussian offset learning and feature refinement that sharpen high-frequency details and stabilize reconstruction. Extensive experiments demonstrate that SR3R outperforms state-of-the-art (SOTA) 3DSR methods and achieves strong zero-shot generalization, even surpassing per-scene optimization baselines on unseen scenes. 

The main contributions are as follows.
\begin{itemize}
    \item \textbf{A novel formulation of 3DSR}. We reformulate 3DSR as a direct feed-forward mapping from LR views to HR 3DGS representations, eliminating the need for 2DSR pseudo-supervision and per-scene optimization. This shifts 3DSR from a 3DGS self-optimization paradigm to a generalized, feed-forward prediction.

    \item  \textbf{A plug-and-play feed-forward framework for sparse-view 3DSR}. We propose SR3R, a feed-forward framework that directly reconstructs HR 3DGS from as few as two LR views through a learned mapping network. SR3R is plug-and-play with any feed-forward 3DGS reconstruction backbone and supports scalable cross-scene training.
    %    cross-scene data to train a generalized 3DSR model.

    \item  \textbf{Gaussian offset learning with feature refinement}. We propose learning Gaussian offsets instead of directly regressing HR Gaussian parameters, which improves learning stability and reconstruction fidelity. In addition, we incorporate a feature refinement to further enhance high-frequency texture details.
    %We design a resolution-mapping network equipped with feature correction and offset learning modules, which effectively enhance high-frequency geometry and texture details while stabilizing the mapping.

    \item \textbf{SOTA performance and robust generalization}. Extensive experiments on three 3D benchmarks demonstrate that SR3R surpasses SOTA 3DSR methods and exhibits strong zero-shot generalization, even outperforming per-scene optimization baselines on unseen scenes.
    %. Notably, SR3R demonstrates strong zero-shot generalization and outperforms per-scene optimization baselines on unseen scenes.
\end{itemize}
\section{Related Work}
\label{sec:related}

\begin{figure*}[htbp]
\centering
\vspace{-10pt}
\includegraphics[width=1\linewidth]{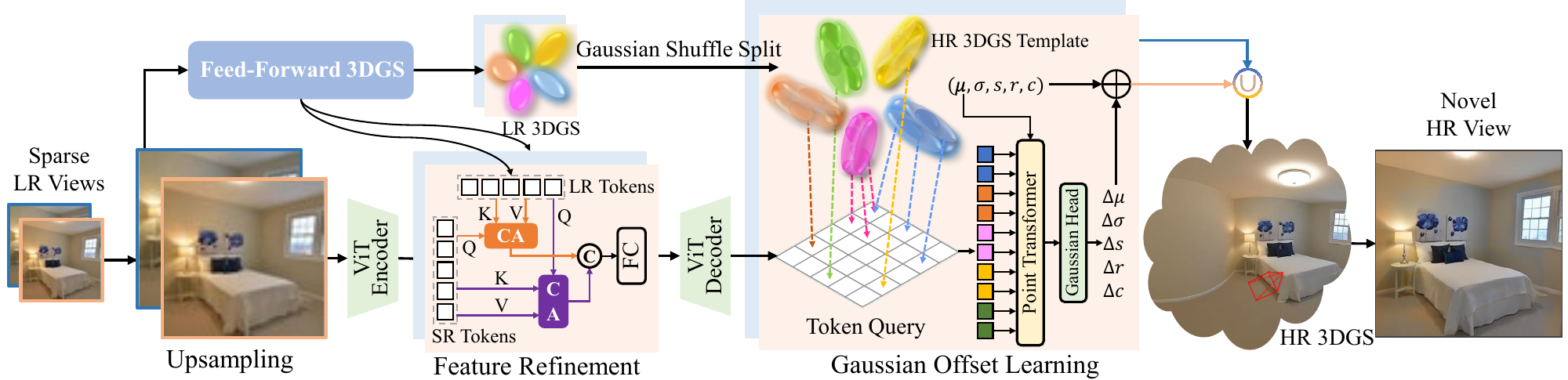}
\caption{\textbf{Overview of the SR3R framework.} Given two LR input views, a feed-forward 3DGS backbone produces an LR 3DGS, which is then densified via Gaussian Shuffle Split to form a structural scaffold. The LR views are upsampled and processed by our mapping network: a ViT encoder with feature refinement integrates LR 3DGS-aware cues, and a ViT decoder performs cross-view fusion. The Gaussian offset learning module then predicts residual offsets to the dense scaffold, yielding the final HR 3DGS for high-fidelity rendering.}
\label{fig:pipeline}
\vspace{-10pt}
\end{figure*}

\subsection{3D Reconstruction}
3DGS \cite{3Dgaussians} has shown remarkable success in 3D scene reconstruction, offering real-time, high-fidelity rendering via Gaussian representations \cite{MipSplatting,gaussianpro,analytic}. However, standard 3DGS reconstruction pipelines rely on dense multi-view inputs \cite{shi2025mmgs,corgs} and per-scene optimization \cite{pixelsplat}, severely limiting their scalability and applicability in real-time or open-world settings. To overcome these constraints, feed-forward 3DGS \cite{pixelsplat,depthsplat,chen2024mvsplat,weng2025gaussianlens} reconstruction models directly infer Gaussian parameters from input views using neural networks, enabling fast, end-to-end reconstruction. Recent extensions have even removed the need for known camera poses \cite{nopo}, further improving their practicality. This framework has been gradually applied in fields such as stylization \cite{wang2025styl3r,liu2025stylos} and scene understanding \cite{xu2025siu3r}. Despite these advances, the current 3D reconstruction quality remains highly sensitive to input image resolution, resulting in significant loss of geometric and texture details under LR conditions. Our proposed SR3R addresses this challenge, enabling high-quality 3D reconstruction from as few as two LR views in a fully feed-forward manner.
%Our proposed SR3R, a feed-forward framework that can achieve high-quality 3D reconstruction from as few as two low-resolution views, addresses this challenge.

\subsection{2D Super-Resolution}
2DSR aims to reconstruct HR images or video frames from their LR counterparts by learning an LR-to-HR image mapping. Over the past decade, the field has seen significant advances driven by model architectures, evolving from early convolutional networks \cite{FSRCNN,EDSR,rcan,basicvsr} to transformer-based architectures \cite{Swinir,EDT} and, more recently, to generative approaches based on adversarial \cite{SRGAN,ESRGAN,xu2024videogigagan} and diffusion models \cite{SR3,resshift,FlashVSR,idm}. The availability of large-scale datasets has further fueled the success of 2DSR. However, 2DSR models face fundamental limitations when applied to 3D scene reconstruction. Since they operate solely in the image domain, they cannot enforce cross-view consistency \cite{feng2024srgs}, often leading to texture artifacts and geometric ambiguity when used to supervise 3D representations. Moreover, domain gaps between natural 2D images and multi-view 3D data further reduce the reliability of 2DSR priors. These limitations raise a central question: instead of relying on 2DSR, can we learn a direct mapping from LR views to HR 3D scene representations? This motivates us to propose SR3R, which directly addresses this problem.

\subsection{3D Super-Resolution}
3DSR aims to reconstruct HR 3D scene representations from LR multi-view images \cite{sequence,CROC}. Recent 3DGS-based 3DSR methods \cite{feng2024srgs,supergs,SuperGaussian,sequence,gaussiansr} address this by injecting high-frequency information derived from pretrained 2DSR models. Typically, pseudo-HR images are generated from dense multi-view LR inputs to supervise the self-optimization of HR 3DGS, while additional regularization, such as confidence-guided fusion \cite{supergs} or radiance field correction \cite{feng2024srgs}, is applied to reduce view inconsistency caused by 2D pseudo-supervision. However, these pipelines suffer from critical limitations. Reconstruction fidelity is bounded by the quality of pseudo-HR labels, and per-scene optimization is computationally expensive and prevents cross-scene learning, limiting scalability. Inspired by recent advances in feed-forward 3DGS reconstruction, we propose SR3R, a feed-forward 3DSR framework that directly maps from LR views to HR 3DGS representations, enabling high-quality 3D reconstruction from as few as two LR input views while supporting efficient, cross-scene generalization.

\section{Methodology}

\subsection{Problem Formulation}
We reformulate 3DGS-based 3DSR as a feed-forward mapping problem from LR multi-view images to an HR 3DGS representation. Unlike prior methods that rely on dense inputs and per-scene optimization supervised by pseudo-HR 2D labels, our formulation enables direct HR 3DGS reconstruction from as few as two LR views, without any per-scene optimization. This removes the reliance on 2DSR pseudo-supervision, allows learning from large-scale multi-scene data, and enables cross-scene generalization, substantially improving scalability and efficiency.

Formally, given a set of $V$ LR input views with camera intrinsics $\{(\boldsymbol{I}^{v}_{lr}, \boldsymbol{K}^{v})\}_{v=1}^{V}$, our goal is to learn a feed-forward mapping function $f_{\boldsymbol{\theta}}$ that predicts an HR 3DGS representation $\mathcal{G}^{\text{HR}}$. Each 3D Gaussian primitive is parameterized by its center $\boldsymbol{\mu}\in\mathbb{R}^3$, opacity $\alpha\in\mathbb{R}$, quaternion rotation $\boldsymbol{r}\in\mathbb{R}^4$, scale $\boldsymbol{s}\in\mathbb{R}^3$, and spherical harmonics (SH) appearance coefficients $\boldsymbol{c}\in\mathbb{R}^{k}$, where $k$ is the number of SH components. For simplicity, we omit the superscript for all Gaussian parameters. The mapping is defined as:
\begin{equation}
f_{\boldsymbol{\theta}} :\left\{\left(\boldsymbol{I}^v_{lr},\, \boldsymbol{K}^v\right)\right\}_{v=1}^{V} \mapsto \mathcal{G}^{\text{HR}} 
\label{eq:mapping}
\end{equation}
where $\mathcal{G}^{\text{HR}} = \{\cup \left(\boldsymbol{\mu}_i^{v},\, \boldsymbol{\alpha}_i^{v},\, \boldsymbol{r}_i^{v},\, \boldsymbol{s}_i^{v},\, \boldsymbol{c}_i^{v}\right) \}_{i=1,\dots,N}^{v=1,\dots,V}$, $\boldsymbol{\theta}$ denotes the learnable parameters of the neural network, and $N$ is the number of Gaussian primitives in $\mathcal{G}^{\text{HR}}$. We omit the view index $v$ hereafter for brevity. 

% \begin{equation}
%     f_{\boldsymbol{\theta}}: \{\left(\boldsymbol{I}^v_{lr}, \boldsymbol{K}^v\right)\}_{v=1}^V \mapsto \mathcal{G}^{HR} = \{\cup \left(\boldsymbol{\mu}_j^v, \boldsymbol{\alpha}_j^v, \boldsymbol{r}_j^v, \boldsymbol{s}_j^v, \boldsymbol{c}_j^v\right)\}_{j=1,\dots,N}^{v=1,\dots,V}
% \end{equation}

\subsection{Overall Framework}
An overview of the proposed SR3R framework is illustrated in Figure \ref{fig:pipeline}. Given two LR input views, SR3R first reconstructs their LR 3DGSs $\mathcal{G}^{\text{LR}}$ using any pretrained feed-forward 3DGS reconstruction model, highlighting the plug-and-play nature of our design. Each $\mathcal{G}^{\text{LR}}$ is then densified via a Gaussian Shuffle Split operation \cite{wan2025s2gaussian} to produce $\mathcal{G}^{\text{Dense}}$, which provides a structural scaffold for high-frequency geometry and texture recovery.
%in the subsequent HR reconstruction.

The LR input images are upsampled to the target resolution and processed by our mapping network, which consists of a ViT encoder, a feature refinement module, a ViT decoder, and a Gaussian offset learning module. The ViT encoder extracts mid-level feature tokens $\boldsymbol{t}_{\text{en}}$, which are refined through cross-attention with intermediate features from the feed-forward 3DGS backbone to produce corrected feature tokens $\boldsymbol{t}_{\text{ca}}$. The ViT decoder then performs cross-view fusion to generate $\boldsymbol{t}_{\text{de}}$, integrating complementary information from both views and mitigating misalignment or ghosting caused by pose inaccuracies or limited overlap. Finally, the Gaussian offset learning module predicts residual offsets from $\mathcal{G}^{\text{Dense}}$ to the target HR 3DGS $\mathcal{G}^{\text{HR}}$. Learning offsets rather than directly regressing HR Gaussian parameters yields more stable training and significantly improves high-frequency texture fidelity, substantially enhancing overall reconstruction quality (Table \ref{tab:real_world}).

%%Specifically, each upsampled image and its camera intrinsics are first processed by the ViT encoder to produce mid-level feature tokens $\boldsymbol{t}_{\text{en}}$. These tokens $\boldsymbol{t}_{\text{en}}$ are refined via cross-attention with intermediate features from the pretrained 3DGS reconstruction backbone, generating corrected tokens $\boldsymbol{t}_{\text{ca}}$. The refined features are then fused via the ViT decoder with cross-view attention to yield $\boldsymbol{t}_{\text{de}}$, which integrates information across both views and alleviates misalignment and ghosting caused by pose inaccuracies or limited overlap. This design maintains geometric consistency across views under minimal geometric priors, reducing dependency on view overlap and enabling cross-view interaction. Finally, the feature offset learning module predicts a residual offset field between the dense Gaussians $\mathcal{G}^{\text{Dense}}$ and the desired HR 3DGS, generating the final output $\mathcal{G}^{\text{HR}}$. Instead of directly regressing the HR Gaussian parameters, SR3R is designed to learn the offset relationship, which proves more efficient and robust. As shown in \autoref{tab:real_world}, this design effectively improves high-frequency texture fidelity and overall reconstruction quality.

%\subsection{LR 3DGS Construction and Gaussian Shuffle Split}
\subsection{LR 3DGS Reconstruction and Densification}
LR 3DGS $\mathcal{G}^{\text{LR}}$ for each input LR view can be obtained by any feed-forward 3DGS model. We then densify them via the Gaussian Shuffle Split operation \cite{wan2025s2gaussian} to produce $\mathcal{G}^{\text{Dense}}$, which serves as a finer structural scaffold for capturing high-frequency geometry and texture details and forms the basis for subsequent Gaussian offset learning.

Each Gaussian primitive $G_j^{\text{LR}}\!=\!(\boldsymbol{\mu}_j,\, \boldsymbol{\alpha}_j,\, \boldsymbol{r}_j,\, \boldsymbol{s}_j,\, \boldsymbol{c}_j)$ in $\mathcal{G}^{\text{LR}}$ is replaced by six smaller sub-Gaussians distributed along the positive and negative directions of its three principal axes. The sub-Gaussian centers are shifted from $\boldsymbol{\mu}_j$ by offsets proportional to the scale $\boldsymbol{s}_j=[s_{j,1},s_{j,2},s_{j,3}]$, controlled by a factor $\beta$ (set to 0.5 by default):
\begin{equation}
\boldsymbol{\mu}_{j,k} = \boldsymbol{\mu}_j + \beta\, R_j \, \boldsymbol{e}_k \odot \boldsymbol{s}_j, \quad k=1,\dots,6,
\label{eq:shuffle}
\end{equation}
where $R_j$ is the rotation matrix derived from the quaternion $\boldsymbol{r}_j$, and $\boldsymbol{e}_k$ denotes the unit direction vectors along each positive and negative principal axis. Each sub-Gaussian inherits $\boldsymbol{r}_j$, $\boldsymbol{\alpha}_j$, and $\boldsymbol{c}_j$ from the original, while its scale along the offset axis is reduced to $\tfrac{1}{4}$ of its original to preserve spatial coverage. For stability, this operation is applied only to Gaussians with opacity above 0.5, focusing densification on structurally significant regions. The final densified 3DGS is obtained by aggregating all sub-Gaussians:
\begin{equation}
\mathcal{G}^{\text{Dense}} =
\bigcup_{j=1}^{M}\,\bigcup_{k=1}^{6}
G_{j,k}^{\text{Dense}}, ~~%\quad
G_{j,k}^{\text{Dense}} =
(\boldsymbol{\mu}_{j,k},\, 
\boldsymbol{\alpha}_j,\,
\boldsymbol{r}_j,\,
\boldsymbol{s}_{j,k},\,
\boldsymbol{c}_j),
\label{eq:dense_final}
\end{equation}
where $M$ is the number of Gaussian primitives in $\mathcal{G}^{\text{LR}}$, and $\mathcal{G}^{\text{Dense}}$ contains $N = 6M$ primitives after densification.
%This densification strategy effectively transforms the coarse $\mathcal{G}^{LR}$ into a compact yet expressive $\mathcal{G}^{Dense}$, providing sufficient geometric granularity for high-resolution 3D reconstruction.

\subsection{LR Image to HR 3DGS Mapping}
The mapping network is the core of SR3R, learning a view-consistent transformation from LR input images to feature representations used for HR 3DGS reconstruction. It adopts a transformer-based architecture composed of a ViT encoder, a feature refinement module, a ViT decoder, and a Gaussian offset learning module. This design enables a view-aware mapping from the 2D LR image domain to the 3D Gaussian domain and leverages large-scale multi-scene training to achieve strong cross-scene generalization.
%The mapping network forms the core of SR3R, responsible for transforming the LR input views into view-consistent, high-frequency feature representations that construct HR 3DGS. The network adopts a transformer-based architecture composed of a ViT encoder, a feature correction module, a ViT decoder, and a feature offset learning module. This design enables effective learning of a view-aware mapping between the 2D LR image domain and the 3D Gaussian representation domain, leveraging large-scale multi-scene data to achieve strong cross-scene generalization.

\textbf{ViT Encoder.}
Each input LR image is first upsampled to the target resolution and, together with its camera intrinsics, is projected into a sequence of patch embeddings before being processed by the ViT encoder to produce mid-level feature tokens $\boldsymbol{t}_{\text{en}}$. The encoder learns locally contextualized representations capturing essential texture and geometric cues. Trained across diverse scenes, these tokens remain reasonably aligned across views with minimal geometric priors, facilitating subsequent cross-view fusion.
%%Since the encoder is trained jointly across multiple scenes, the resulting tokens are well-aligned across views under minimal geometric priors, facilitating subsequent cross-view reasoning.
%$\boldsymbol{t}_{en} \in \mathbb{R}^{H' \times W' \times C}$, where $H'$ and $W'$ denote the dimensions of the token grid and $C$ is the feature embedding dimension

\begin{figure*}[t]
    \centering
    \includegraphics[width=0.95\linewidth]{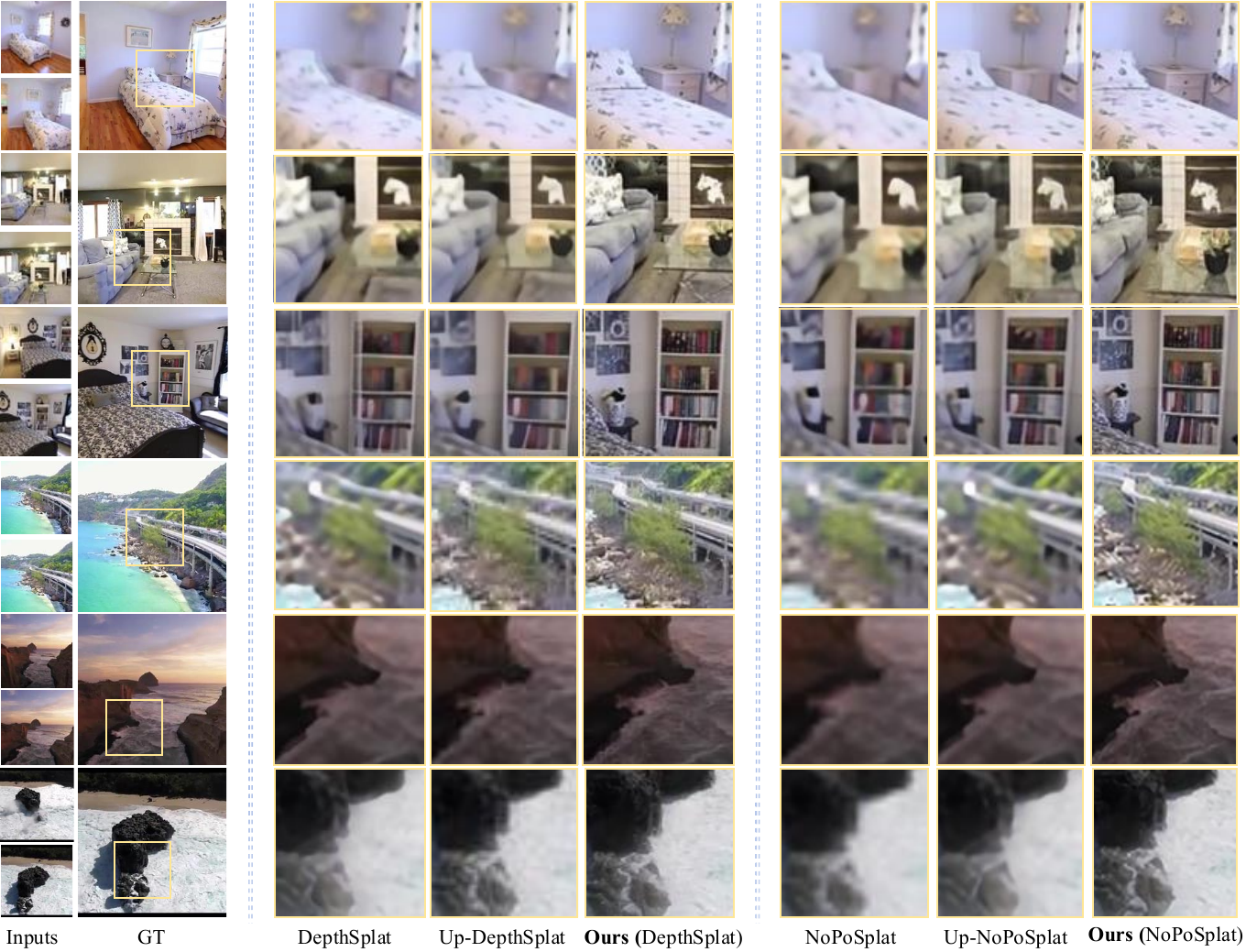}
    \caption{\textbf{Qualitative comparison with SOTA feed-forward 3DGS reconstruction methods on Re10k (top three) and ACID (bottom three) datasets.} SR3R delivers significantly sharper details and more stable geometry than DepthSplat, NoPoSplat, and their upsampled variants, consistently improving reconstruction quality across different 3DGS backbones under sparse LR inputs.}
    \label{fig:main}
\end{figure*}

\textbf{Feature Refinement Module.}
Upsampled LR images often contain ambiguous or hallucinated high-frequency patterns due to interpolation, which may mislead the mapping network and introduce geometric or texture artifacts in 3D. To correct these unreliable 2D features, we introduce a feature refinement module that aligns the encoder tokens $\boldsymbol{t}_{\text{en}}\!\in\!\mathbb{R}^{N \times C}$ with geometry-aware tokens $\boldsymbol{t}_{\text{pre}}\!\in\!\mathbb{R}^{N \times C}$ extracted from the pretrained feed-forward 3DGS backbone used to obtain $\mathcal{G}^{\text{LR}}$. Here, $N$ denotes the number of tokens, and $C$ is the feature embedding dimension. Two cross-attentions are computed in opposite directions:
%%Upsampling LR inputs may introduce aliasing and feature drift, which can propagate into 3D misalignment. To mitigate this, we design a feature correction module that refines the encoder tokens $\boldsymbol{t}_{\text{en}}\!\in\!\mathbb{R}^{N \times C}$ from our network through bidirectional cross-attention with encoder tokens $\boldsymbol{t}_{\text{pre}}\!\in\!\mathbb{R}^{N \times C}$ extracted from the pretrained feed-forward 3DGS backbone used to obtain $\mathcal{G}^{\text{LR}}$. Here, $N$ denotes the number of tokens, and $C$ is the feature embedding dimension. Two cross-attentions are computed in opposite directions:
\begin{equation}
\begin{aligned}
\mathbf{U}_{o\leftarrow p} &= 
\operatorname{softmax}\!\left(
\frac{(\boldsymbol{t}_{\text{en}}\boldsymbol{W}_Q^{o})
(\boldsymbol{t}_{\text{pre}}\boldsymbol{W}_K^{p})^\top}{\sqrt{d}}
\right)(\boldsymbol{t}_{\text{pre}}\boldsymbol{W}_V^{p}), \\[4pt]
\mathbf{U}_{p\leftarrow o} &= 
\operatorname{softmax}\!\left(
\frac{(\boldsymbol{t}_{\text{pre}}\boldsymbol{W}_Q^{p})
(\boldsymbol{t}_{\text{en}}\boldsymbol{W}_K^{o})^\top}{\sqrt{d}}
\right)(\boldsymbol{t}_{\text{en}}\boldsymbol{W}_V^{o}),
\end{aligned}
\label{eq:bi_attention}
\end{equation}
where $o$ and $p$ denote our encoder and the pretrained encoder, respectively, $\boldsymbol{W}_Q^{(\cdot)}$, $\boldsymbol{W}_K^{(\cdot)}$, and $\boldsymbol{W}_V^{(\cdot)}\!\in\!\mathbb{R}^{C\times d}$ are learnable projection matrices, and $d$ is the feature dimension per attention head. The two attention outputs $\mathbf{U}_{o\leftarrow p}$ and $\mathbf{U}_{p\leftarrow o}$ are then concatenated and fused through a fully connected layer to generate the refined feature token $\boldsymbol{t}_{ca}$. This refinement process transfers reliable 3D geometric priors from the pretrained 3DGS encoder into our 2D feature space, suppressing upsampling-induced ambiguities and producing features that are better aligned with the underlying Gaussian structure and more consistent across views.

\textbf{ViT Decoder.}
The refined features $\boldsymbol{t}_{\text{ca}}$ from both views are fed into a ViT decoder, which performs intra-view self-attention to aggregate global contextual information and inter-view cross-attention to fuse cross-view features. This produces the decoded features $\boldsymbol{t}_{\text{de}}\!\in\!\mathbb{R}^{N\times C}$, which integrate multi-view geometry and reduce inconsistencies caused by pose inaccuracy or limited view overlap. The decoded features are then provided to the Gaussian offset learning module (Section~\ref{sec:offset}) to estimate residual corrections from the densified representation $\mathcal{G}^{\text{Dense}}$ to the target HR 3DGS $\mathcal{G}^{\text{HR}}$.
%%for each view. The decoder enhances each view’s representation by capturing both fine-grained local details and global scene-level geometry, thereby reducing pose-related misalignment and multi-view ghosting artifacts. The resulting features $\boldsymbol{t}_{\text{de}}$ are subsequently used by the feature offset learning module (see Section~\ref{sec:offset}) to predict the residual transformation from the densified Gaussian representation $\mathcal{G}^{\text{Dense}}$ to the target HR 3DGS $\mathcal{G}^{\text{HR}}$.

%Overall, the LR-to-HR mapping network establishes a differentiable bridge between the 2D LR image space and the 3D Gaussian domain, producing geometrically consistent, high-frequency representations that enable efficient and robust HR 3D reconstruction.

\begin{table*}[!ht]
    \renewcommand{\arraystretch}{1.2}
    \footnotesize
    \centering
    \caption{\textbf{Quantitative comparison of 4× 3DSR on the large-scale RE10K and ACID datasets}. SR3R consistently and substantially outperforms all baselines and their upscaled-input versions across PSNR, SSIM, and LPIPS, with only moderate Gaussian complexity and training memory. \textbf{Bold} indicates the best results and \underline{underline} the second best.}
    \label{tab:real_world}
    \begin{tabular}{ll@{\hspace{0.5cm}}c@{\hspace{0.5cm}}ccccc}
    \Xhline{1pt}
    \multirow{2}{*}{\textbf{Dataset}} & \multirow{2}{*}{\textbf{Method}} &  \multicolumn{3}{c}{\textbf{Metrics}} & \multirow{2}{*}{\makecell{\textbf{Gaussian}\\ \textbf{Param.} $\downarrow$}} & \multirow{2}{*}{\makecell{\textbf{Gaussian} \\ \textbf{Num.} $\downarrow$}} & \multirow{2}{*}{\makecell{\textbf{Training} \\ \textbf{Mem.} $\downarrow$}} \\
    \cline{3-5}
    & & PSNR$\uparrow$ & SSIM$\uparrow$ & LPIPS$\downarrow$ & &  & \\
    \Xhline{1pt}
    \multirow{6}{*}{\makecell[l]{RE10K\\ 64$\times$64 $\rightarrow$ 256$\times$256}} 
    & NoPoSplat \cite{Ye2025} & 21.326 & 0.612 & 0.307 & 2.7M & 8,192 & 4.82GB \\
    & Up-NoPoSplat & 23.374 & 0.771 & 0.251 & 44.5M & 131,072 & 21.36GB \\
    & \textbf{Ours (NoPoSplat)} & \textbf{24.794} & \textbf{0.827} & \textbf{0.188} & \underline{16.5M} & \underline{49,152} & \underline{12.92GB}  \\ 
    \cline{2-8}
    & DepthSplat \cite{xu2025depthsplat} & 23.147 & 0.699 & 0.281 & 2.3M & 8,192 & 7.25GB \\
    & Up-DepthSplat & 24.712 & 0.793 & 0.244 & 38.3M & 131,072 & 26.17GB \\
    & \textbf{Ours (DepthSplat)} & \textbf{26.250} & \textbf{0.856} & \textbf{0.165} & \underline{14.2M} & \underline{49,152} & \underline{17.43GB} \\
    \Xhline{1pt}
    \multirow{6}{*}{\makecell[l]{ACID\\ 64$\times$64 $\rightarrow$ 256$\times$256}} 
    & NoPoSplat \cite{Ye2025} & 21.451 & 0.606 & 0.531 & 2.7M & 8,192 & 4.82GB \\
    & Up-NoPoSplat & 23.911 & 0.692 & 0.384 & 44.5M & 131,072 & 21.36GB \\
    & \textbf{Ours (NoPoSplat)} & \textbf{25.541} & \textbf{0.746} & \textbf{0.283} & \underline{16.5M} & \underline{49,152} & \underline{12.92GB} \\ 
    \cline{2-8}
    & DepthSplat \cite{xu2025depthsplat} & 23.801 & 0.624 & 0.437 & 2.3M & 8,192 & 7.25GB \\
    & Up-DepthSplat & 25.315 & 0.721 & 0.322 & 38.3M & 131,072 & 26.17GB \\
    & \textbf{Ours (DepthSplat)} & \textbf{27.018} & \textbf{0.797} & \textbf{0.261} & \underline{14.2M} & \underline{49,152} & \underline{17.43GB} \\
    \Xhline{1pt}
    \end{tabular}
\end{table*}

\subsection{Gaussian Offset Learning} \label{sec:offset}
%Given the highly non-linear and scene-dependent relationship between 2D appearance and 3D geometry, directly regressing absolute Gaussian parameters from image features is often inefficient and unstable. Instead, we proposed to learn a \textbf{Gaussian offset field} that refines the densified Gaussian representation $\mathcal{G}^{\text{Dense}}$ into the target HR representation $\mathcal{G}^{\text{HR}}$. This design improves robustness (\autoref{tab:real_world}) by constraining the learning objective to local geometric and photometric offset rather than full parameter regression.
Given the non-linear and scene-dependent relationship between 2D appearance and 3D geometry, directly regressing absolute Gaussian parameters from image features is often inefficient and unstable, as the resulting prediction space is large and multi-modal. In contrast, the densified representation $\mathcal{G}^{\text{Dense}}$ already provides a reliable structural scaffold, meaning that the remaining discrepancy to HR is primarily local and high-frequency. Motivated by this, we proposed to learn a \textbf{Gaussian offset field} that predicts residual corrections to $\mathcal{G}^{\text{Dense}}$ rather than regressing full HR parameters. This formulation constrains the learning target to local geometric and photometric offset, leading to more stable optimization and sharper reconstruction quality (Table \ref{tab:real_world}).

Specifically, for each Gaussian primitive $G_i^{\text{Dense}}\!=\!(\boldsymbol{\mu}_i,\, \boldsymbol{\alpha}_i,\, \boldsymbol{r}_i,\, \boldsymbol{s}_i,\, \boldsymbol{c}_i)$ in $\mathcal{G}^{\text{Dense}}$, we project its 3D center $\boldsymbol{\mu}_i$ onto the image plane to obtain the 2D coordinate $\boldsymbol{p}_i$. The corresponding local feature $\boldsymbol{F}_i$ is then sampled from the reshaped decoded feature map $\boldsymbol{t}_{de}$ at location $\boldsymbol{p}_i$'s patch. These queried features are aggregated together with the Gaussian center and camera intrinsics $\boldsymbol{K}$, and passed into a PointTransformerV3 network for spatial reasoning and multi-scale feature encoding:
\begin{equation}
\boldsymbol{F} = \Phi_{\text{PTv3}}\!\left(\left[\boldsymbol{\mu}_i;\, \{\boldsymbol{F}_i\}_{i=1}^N;\, \boldsymbol{K} \right]\right),
\label{eq:ptv3}
\end{equation}
where $\Phi_{\text{PTv3}}$ denotes the PointTransformerV3 encoder that captures geometric relations and contextual dependencies among neighboring Gaussians. The encoded feature $\boldsymbol{F}$ is then fed into a Gaussian Head $\Psi_{\text{GH}}$, a lightweight MLP that predicts residual offsets for the Gaussian parameters:
\begin{equation}
\Delta G = (\Delta\boldsymbol{\mu},\, \Delta\boldsymbol{\alpha},\, 
\Delta\boldsymbol{r},\, \Delta\boldsymbol{s},\, \Delta\boldsymbol{c}) 
= \Psi_{\text{GH}}(\boldsymbol{F}).
\label{eq:offset}
\end{equation}
%Here, $\Delta G$ denotes the residual offsets for individual Gaussians, and the complete offset field is represented as $\Delta\mathcal{G}={\Delta G_j}_{j=1}^{N}$. 
The final HR 3DGS is obtained via residual composition:
\begin{equation}
\mathcal{G}^{\text{HR}} = \mathcal{G}^{\text{Dense}} + \Delta\mathcal{G}, ~~~~ \Delta\mathcal{G}={\Delta G_i}_{i=1}^{N}
\end{equation}
This residual formulation naturally focuses the network on high-frequency refinements while preserving the coarse structure encoded by $\mathcal{G}^{\text{Dense}}$. Compared with direct parameter regression, it improves convergence stability, reduces artifacts, and consistently yields sharper textures and more accurate geometry.
%%This design enables the network to focus on learning high-frequency geometric and appearance refinements while preserving the structural scaffold provided by $\mathcal{G}^{\text{Dense}}$. Compared with direct parameter regression, the offset learning paradigm improves convergence stability, reduces visual artifacts, and ensures that the reconstructed $\mathcal{G}^{\text{HR}}$ achieves superior texture fidelity and geometric consistency.

\subsection{Training Objective}
The predicted HR 3DGS $\mathcal{G}^{\text{HR}}$ is rendered into novel-view images and supervised using the corresponding ground-truth RGB observations. The entire SR3R is trained end-to-end through differentiable Gaussian rasterization. Following \cite{Ye2025}, we adopt a combination of pixel-wise reconstruction loss (MSE) and perceptual consistency loss (LPIPS) to jointly preserve geometric accuracy and visual fidelity.
%The predicted HR 3DGS $\mathcal{G}^{\text{HR}}$ is rendered into 2D images at novel viewpoints, with the rendered results supervised by corresponding ground-truth RGB images. The entire SR3R framework is trained end-to-end, where gradients are back-propagated through the differentiable Gaussian rasterization process to optimize the mapping network. Following \cite{Ye2025}, we employ the same combination of pixel-wise reconstruction loss (MSE) and perceptual consistency loss (LPIPS) to ensure geometric accuracy and visual fidelity.

\section{Experimental Results}

\begin{figure*}[htbp]
    \centering
    \includegraphics[width=0.83\linewidth]{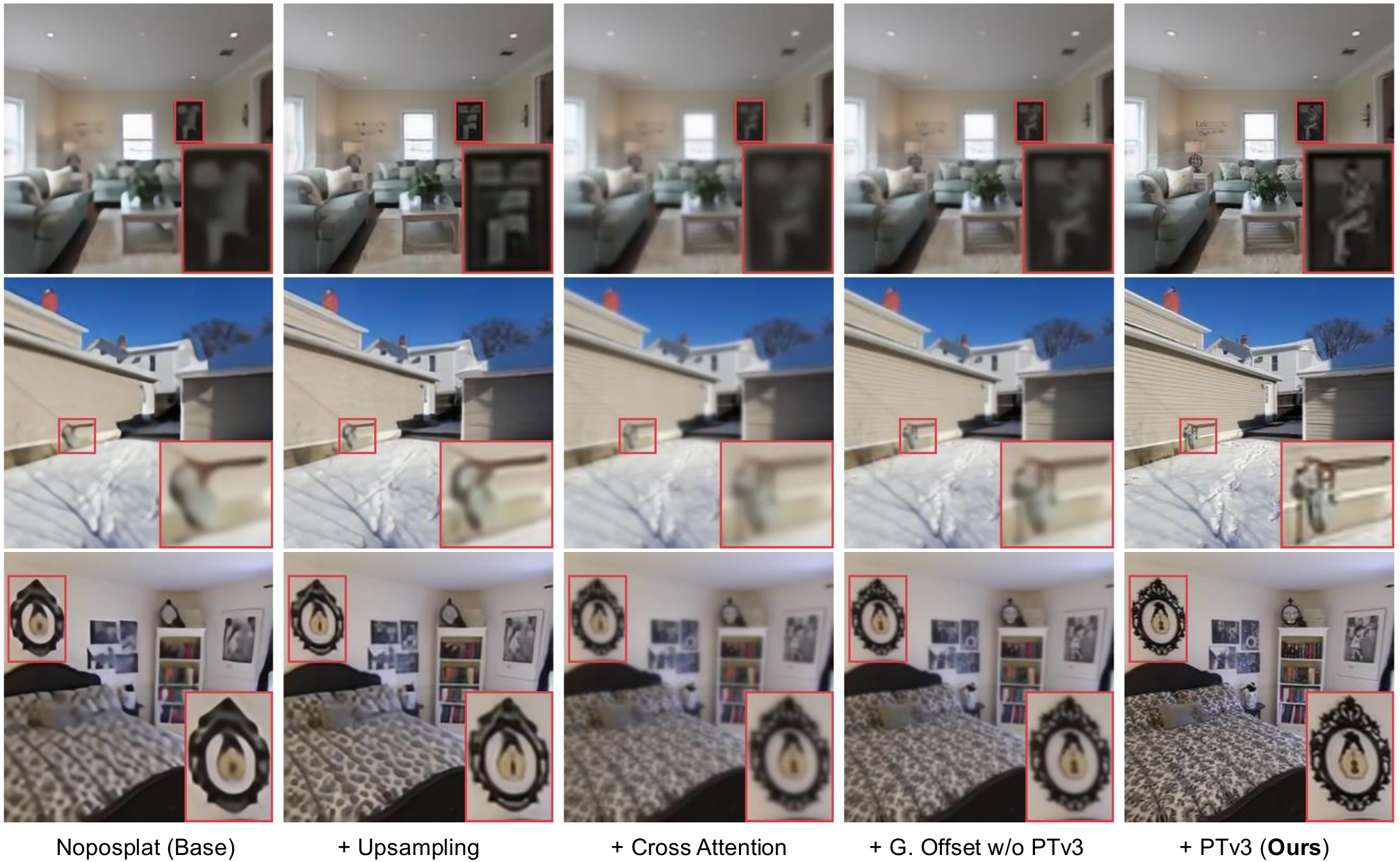}
    \caption{\textbf{Qualitative ablation results of SR3R components.} Each component of SR3R progressively improves reconstruction quality, with upsampling reducing coarse blur, cross-attention improving feature alignment, Gaussian offset learning enhancing local geometry, and PTv3 yielding the sharpest and most consistent results.}
    \label{fig:ablation}
\end{figure*}

\subsection{Experimental Setup}
\textbf{Datasets.}
We evaluate SR3R on three widely used 3D datasets: RealEstate10K (RE10K) \cite{Re10k}, ACID \cite{ACID}, and DTU \cite{DTU}. RE10K and ACID are two large-scale datasets, containing indoor real estate walkthrough videos and outdoor natural scenes captured by aerial drones, respectively. For fair comparison, we follow the official train–test splits used in prior works \cite{nopo,depthsplat}. To further assess generalization, we perform zero-shot 3DSR experiments on the DTU dataset, which features object-centric scenes with different camera motion and scene types from the RE10K.
%with diverse geometries and illumination conditions.
%, and DL3DV \cite{}, DL3DV provides large-scale, high-resolution 4K multi-view sequences covering diverse indoor and outdoor environments, offering richer scene complexity for generalization analysis. 

%\noindent 
\textbf{Baselines and Metrics.}
We compare SR3R with two state-of-the-art feed-forward 3DGS reconstruction models, NoPoSplat \cite{nopo} and DepthSplat \cite{depthsplat}, as well as the per-scene optimization methods SRGS \cite{feng2024srgs} and FSGS \cite{zhu2024fsgs}. This setup allows us to evaluate large-scale 3DSR performance and demonstrate SR3R’s superior zero-shot capability without scene-specific optimization. Following prior work \cite{feng2024srgs,sequence}, we assess novel-view synthesis quality using PSNR, SSIM, and LPIPS \cite{lpip}.
%%We compare SR3R with two SOTA feed-forward 3DGS reconstruction models, NoPoSplat \cite{nopo} and DepthSplat \cite{depthsplat}, to evaluate large-scale 3DSR performance. We further include the per-scene optimization methods SRGS \cite{feng2024srgs} and FSGS \cite{zhu2024fsgs} as baselines to demonstrate that SR3R achieves superior performance even under a zero-shot setting without any scene-specific optimization. Following prior works \cite{feng2024srgs,sequence}, we evaluate 3DSR performance via novel view synthesis quality using PSNR, SSIM, and LPIPS \cite{lpip}.

% \textbf{Evaluation Metrics.}
% Following prior works \cite{}, we evaluate 3DSR performance via novel view synthesis quality using PSNR, SSIM, and LPIPS.
%which measure pixel-level accuracy, structural similarity, and perceptual fidelity, respectively.

\textbf{Implementation Details.}
We implement SR3R in PyTorch and evaluate its plug-and-play compatibility with two 3DGS reconstruction backbones, NoPoSplat \cite{nopo} and DepthSplat \cite{depthsplat}. Input images are preprocessed by rescaling and center cropping, where the LR inputs are downsampled to $64\times64$ and the ground-truth (GT) targets to $256\times256$ using the LANCZO resampling filter. SwinIR \cite{Swinir} is used as the upsampling backbone, while simpler operators such as Bicubic yield comparable results (Table \ref{tab:ablation2}). The ViT encoder–decoder follows a vanilla configuration with a patch size of 16 and 8 attention heads. The MSE and LPIPS loss weights follow \cite{nopo} and are set to 1 and 0.05. Both the backbone and our mapping network are trained for 75{,}000 iterations with a batch size of 8 and a learning rate of $2.5\times10^{-5}$. All experiments are conducted on four NVIDIA RTX 5090 GPUs.

\subsection{Comparison with State-of-the-Art}
We evaluate SR3R through 4$\times$ 3DSR experiments on the large-scale RE10K and ACID datasets, and compare it against the SOTA feed-forward 3DGS reconstruction models NoPoSplat and DepthSplat. In addition to their standard version, we further evaluate their upsampled-input variants (\emph{Up-NoPoSplat} and \emph{Up-DepthSplat}), where LR inputs are first upsampled before direct HR Gaussian regression.
%being fed into the reconstruction networks to directly regress HR Gaussian parameters.

%%Table \ref{tab:real_world} reports the comparison results. SR3R consistently and substantially outperforms both original and upsampled-input baselines across all evaluation metrics on both datasets, demonstrating its superior capability in recovering fine-grained geometry and texture details. These results further validate the advantage of our proposed Gaussian offset learning, which achieves higher-fidelity HR 3D reconstructions compared to direct parameter regression. We also report model complexity and training cost, showing that SR3R achieves these substantial gains with moderate computational overhead, underscoring its practicality for scalable feed-forward 3DSR.
Table \ref{tab:real_world} shows that SR3R consistently outperforms both original and upsampled-input baselines across all metrics on both datasets. These results highlight the advantage of learning Gaussian offsets over direct parameter regression, enabling more accurate high-frequency recovery under sparse LR inputs. We also report complexity and training cost, showing that SR3R achieves these substantial gains with moderate computational overhead, demonstrating its practicality for scalable feed-forward 3DSR.

Figure \ref{fig:main} provides qualitative comparisons. Both baselines exhibit blurring, texture flattening, and geometric instability, while their upsampled variants remain unable to recover reliable high-frequency details and often introduce hallucinated edges or ghosting artifacts. In contrast, SR3R reconstructs sharper textures, cleaner boundaries, and more consistent geometry across views. These improvements hold for both 3DGS backbones, confirming that our offset-based refinement and cross-view fusion effectively restore 3D-specific high-frequency structures that 2D upsampling and direct HR regression cannot recover.

%%Both feed-forward baselines exhibit severe blurring, texture flattening, and geometric inconsistencies. Their upsampled variants alleviate some degradation but still fail to reconstruct reliable high-frequency details, often introducing hallucinated edges or residual ghosting. In contrast, SR3R consistently produces sharper textures, clearer object boundaries, and more stable geometry across views. These improvements hold for both 3DGS backbones, demonstrating that our offset-based refinement and cross-view feature fusion effectively recover 3D-specific high-frequency structures that cannot be restored by full HR 3DGS parameter regression with 2D upsampling.

\subsection{Zero-Shot Generalization}
We further evaluate the zero-shot generalization ability of SR3R on the DTU dataset, a challenging object-centric benchmark with unseen geometries and illumination conditions. All feed-forward models, including SR3R and baselines, are trained on RE10K and directly tested on DTU without any fine-tuning. We additionally include two SOTA per-scene optimization methods, SRGS \cite{feng2024srgs} and FSGS \cite{zhu2024fsgs}, a sparse-view-specific model that we combine with SRGS (denoted as FSGS+SRGS) to provide a stronger baseline.

As shown in Table \ref{tab:zero-shot}, SR3R achieves substantially higher accuracy than all feed-forward baselines in the zero-shot setting, demonstrating strong cross-scene generalization. Notably, SR3R also surpasses the per-scene optimization methods SRGS and FSGS+SRGS, despite requiring no scene-specific fitting at test time. This indicates that SR3R effectively preserves geometric and photometric fidelity even on completely unseen scenes. In terms of efficiency, SR3R is significantly faster than optimization-based methods, enabling practical real-time inference. Although its inference cost is slightly higher than that of other feed-forward models, the clear performance gains make SR3R a compelling choice for scalable 3DSR.

%%Table \ref{tab:zero-shot} summarizes the comparison. SR3R achieves substantially higher reconstruction accuracy than all feed-forward baselines in the zero-shot setting, demonstrating strong cross-scene generalization. Notably, SR3R also surpasses the per-scene optimization methods SRGS and FSGS+SRGS, even though they rely on scene-specific optimization while SR3R performs no fitting at test time. This highlights the robustness of our method in preserving high-frequency geometry and appearance on completely unseen scenes. In terms of efficiency, SR3R is orders of magnitude faster than optimization-based methods, enabling practical real-time inference. Although SR3R incurs slightly higher inference cost than other feed-forward baselines, the significant performance gains fully justify the overhead.

\begin{table}[!t]
    \renewcommand{\arraystretch}{1.2}
    \footnotesize
    \centering
    \caption{\textbf{Zero-shot generalization results from RE10K to DTU.} Feed-forward models are trained on RE10K and tested on DTU without fine-tuning. SRGS and FSGS+SRGS use per-scene optimization. SR3R delivers the best reconstruction quality while remaining significantly faster than optimization-based methods. \textbf{Bold} indicates the best results and \underline{underline} the second best.}
    \label{tab:zero-shot}
    \resizebox{\linewidth}{!}
    {
    \begin{tabular}{l cccc}
    \Xhline{1pt}
    \multirow{2}{*}{\textbf{Method}} & \multicolumn{4}{c}{\textbf{RE10K $\rightarrow$ DTU}}  \\
    \cline{2-5} 
          & PSNR $\uparrow$   & SSIM $\uparrow$  & LPIPS $\downarrow$ & Rec. Time $\downarrow$  \\ 
    \Xhline{1pt}
    SRGS \cite{feng2024srgs}   &12.420 &0.327 &0.598 & 300s\\
    FSGS+SRGS \cite{zhu2024fsgs} &13.720 &0.444 &0.481  &420s\\
    \hline
    NopoSplat \cite{Ye2025}  &12.628 &0.343 &0.581 &0.01s \\ %0.009
    Up-Noposplat &16.643 &0.598 &0.369 &\underline{0.16s}\\ %0.013
    %\hline
    \textbf{Ours (NopoSplat) } &\textbf{17.241} &\textbf{0.607} &\textbf{0.291} & 1.69s\\ %1.699
    \Xhline{1pt}
    \end{tabular}
    }
\end{table}

\subsection{Ablation Study}
\textbf{Component Analysis.}
To assess the contribution of each component in SR3R, we perform a component-wise ablation using NoPoSplat as the baseline and evaluate 4$\times$ 3DSR performance on RE10K. As reported in Table \ref{tab:ablation1}, all proposed modules bring consistent and significant improvements. Adding the upsampling module provides a stronger initial estimate and yields clear improvements. Incorporating bidirectional cross-attention further enhances structural consistency by injecting geometric priors from the pretrained 3DGS encoder. Gaussian Offset Learning yields the largest performance gain. Even without PTv3 (G. Offset w/o PTv3), it significantly improves reconstruction quality while reducing the number of learnable Gaussian parameters, demonstrating its efficiency. Adding PointTransformerV3 further boosts accuracy through multi-scale spatial reasoning, producing the full SR3R model with the best performance. These results confirm that all components are necessary and complementary, collectively enabling SR3R to achieve high-fidelity HR 3D reconstruction.
%%Results in Table \ref{tab:ablation1} show that each proposed module contributes consistently and significantly. Adding the upsampling module provides a stronger initial estimate and yields clear improvements. Incorporating bidirectional cross-attention further enhances structural consistency by injecting geometric priors from pretrained 3DGS features. The proposed Gaussian Offset Learning brings the most notable gain. Even without PTv3 (G. Offset w/o PTv3), it markedly boosts reconstruction quality while reducing learnable Gaussian parameters from 44.5M to 16.5M, demonstrating its efficiency. Adding PointTransformerV3 further improves performance through multi-scale spatial reasoning, producing the full SR3R model with the best accuracy. These results confirm that all components are necessary and complementary, collectively enabling SR3R to achieve high-fidelity HR 3D reconstruction.

Figure \ref{fig:ablation} presents the qualitative ablation results. The NoPoSplat baseline produces severe blurring and geometric degradation under sparse LR inputs. Applying 2D upsampling reduces excessive softness but still fails to recover reliable high-frequency structures, often introducing ambiguous or hallucinated textures. Adding cross-attention feature refinement improves feature alignment across views and suppresses texture drift. Gaussian Offset Learning further sharpens local geometry and appearance, yielding clearer object boundaries and more stable surface details. Integrating PTv3 completes the model and produces the sharpest textures, most accurate geometry, and highest overall fidelity. These results confirm that each SR3R component contributes progressively and that refinement, offset learning, and PTv3 together are essential for high-quality 3DSR.

%we conduct a step-by-step ablation study using NoPoSplat as the baseline and perform 4× 3DSR evaluation on the RE10K dataset. The results are summarized in \autoref{tab:ablation1}. Starting from the baseline NoPoSplat, introducing the image upsampling module substantially improves reconstruction quality by providing a finer initial estimate for the LR-to-HR mapping. Adding the proposed bidirectional cross-attention mechanism further enhances geometric and photometric consistency by effectively transferring structural cues from the pretrained 3DGS features. A particularly notable gain comes from the proposed Gaussian Offset Learning module. Even without the PointTransformerV3 aggregator (G. Offset w/o PTv3), this module leads to a large improvement across all metrics while simultaneously reducing the number of learnable Gaussian parameters from 44.5M to 16.5M, highlighting its efficiency and effectiveness. Incorporating PointTransformerV3 brings an additional performance boost by enabling multi-scale spatial reasoning among Gaussians, yielding the full SR3R model with the best overall accuracy. Overall, each component contributes meaningfully and cumulatively to the final performance, demonstrating the necessity and complementary nature of all modules in SR3R.

\begin{table}[!t]
    \renewcommand{\arraystretch}{1.1}
    \footnotesize
    \centering
    \caption{\textbf{Component-wise ablation on RE10K (4$\times$ 3DSR).} Modules are added cumulatively to the NoPoSplat baseline. Each component improves performance, and Gaussian Offset Learning yields the largest gain with fewer learnable Gaussians. The full SR3R achieves the best results.}
    \label{tab:ablation1}
    \resizebox{\linewidth}{!}
    {
    \begin{tabular}{l cccc}
    \Xhline{1pt}
    \multirow{2}{*}{\textbf{Component}}  & \multicolumn{4}{c}{\textbf{RE10K (64 $\rightarrow$ 256)}}   \\
    \cline{2-5}
         & PSNR $\uparrow$   & SSIM $\uparrow$  & LPIPS $\downarrow$ & \makecell{Gauss. Param. $\downarrow$}   \\ 
    \Xhline{1pt}
    Noposplat(Base)   &21.326 &0.612 &0.307&2.7M \\
    $+$ Upsampling & 23.374 &0.771 &0.251&44.5M \\
    $+$ Cross Attention     &23.504&0.784 &0.237&44.5M \\
    $+$ G. Offset w/o PTv3   &24.447 &0.808 &0.211&16.5M \\
    $+$ PTv3 \textbf{(Ours)} &\textbf{24.794} & \textbf{0.827} & \textbf{0.188} & \underline{16.5M} \\ 
    \Xhline{1pt}
    \end{tabular}
    }
\end{table}

\textbf{Robustness to Upsampling Strategy.}
We evaluate the robustness of SR3R to different upsampling strategies used before the ViT encoder. Four commonly used methods are tested, including two interpolation-based approaches (Bilinear, Bicubic) and two learning-based SR models (SwinIR \cite{Swinir} and HAT \cite{hat}). As shown in Table \ref{tab:ablation2}, SR3R delivers consistently strong performance across all metrics, with only minor variation across different upsampling choices. Notably, even Bilinear interpolation already surpasses all feed-forward baselines (Table \ref{tab:real_world}), indicating that SR3R does not depend on a particular upsampling design. 

%%We further examine the robustness of SR3R to different upsampling strategies used to generate the input to the ViT encoder. Four commonly used methods are tested, including two interpolation-based methods (Bilinear, Bicubic) and two learning-based SR models (SwinIR \cite{Swinir} and HAT \cite{hat}). The results in Table \ref{tab:ablation2} show that SR3R achieves consistently strong performance across all metrics, with only minor variation among different upsampling choices. Remarkably, even Bilinear interpolation already surpasses all feed-forward baselines (\autoref{tab:real_world}), demonstrating that SR3R is inherently robust to the upsampling design. 
% and that the core reconstruction gains stem from our Gaussian offset learning framework rather than the upsampling module.

\begin{table}[!ht]
    \renewcommand{\arraystretch}{1.1}
    \footnotesize
    \centering
    \caption{\textbf{Ablation on upsampling strategies on RE10K (4$\times$ 3DSR).} SR3R maintains consistently strong performance across all interpolation and learning-based upsampling methods.}
    \label{tab:ablation2}
    %\resizebox{\linewidth}{!}
    {
    \begin{tabular}{l cccc}
    \Xhline{1pt}
    \multirow{2}{*}{\textbf{Upsampling}}     & \multicolumn{4}{c}{\textbf{RE10K (64 $\rightarrow$ 256)}}   \\
    \cline{2-5}
     & PSNR $\uparrow$   & SSIM $\uparrow$  & LPIPS $\downarrow$ & Rec. Time $\downarrow$   \\ 
    \Xhline{1pt}
    Bilinear  &24.586 &0.795 &0.204 &1.59s\\ %1.594
    Bicubic &24.663 &0.817 &0.193 &1.53s\\ %1.531
    SwinIR \cite{Swinir}  &24.794&0.827&0.188 &1.69s\\ %1.699
    HAT \cite{hat} &24.782 &0.819 &0.183  &1.75s\\ %1.757
    \Xhline{1pt}
    \end{tabular}
    }
\end{table}

\section{Conclusion}
We reformulate 3DSR as a feed-forward mapping from sparse LR views to HR 3DGS, enabling the learning of 3D-specific high-frequency priors from large-scale multi-scene data. Based on this new paradigm, SR3R combines feature refinement and Gaussian offset learning to achieve high-quality HR reconstruction with strong generalization. Experiments show that SR3R surpasses prior methods and provides an efficient, scalable solution for feed-forward 3DSR.

%We proposed SR3R, a plug-and-play feed-forward framework that achieves high-quality 3DGS SR directly from sparse LR views. SR3R takes advantage of the excellent generalization of feedforward networks to complete the optimization paradigm shift from knowledge distillation to mapping for 3D super-resolution tasks. This gives 3D super-resolution tasks the ability to self-explore high-frequency details, breaking the previous seal. By introducing an LR image to HR 3DGS mapping network and Gaussian offset learning, SR3R reconstructs high-fidelity details without any 3DGS optimization. Extensive experiments demonstrate that SR3R significantly surpasses SOTA baselines while offering strong zero-shot generalization. 

% and practical efficiency. %bidirectional cross-attention
\clearpage
\setcounter{page}{1}
\maketitlesupplementary

\setcounter{section}{0}
\renewcommand{\thesection}{\Alph{section}}

\setcounter{figure}{0}
\renewcommand{\thefigure}{S\arabic{figure}}

\setcounter{table}{0}
\renewcommand{\thetable}{S\arabic{table}}

\section{More Details for Gaussian Offset Learning}
\begin{figure}[t]
    \centering
    \includegraphics[width=1.0\linewidth]{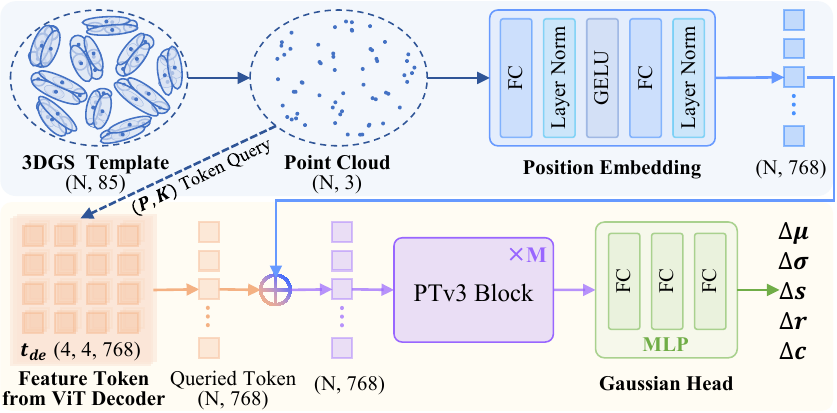}
    \caption{\textbf{Detailed Gaussian Offset Learning pipeline.} Each Gaussian center is projected to the image plane to query local ViT features. The queried token is fused with a geometry-aware position embedding and processed by PTv3 blocks for spatial reasoning. A lightweight Gaussian Head predicts residual offsets to refine the initial 3DGS template.}
    %Each densified Gaussian primitive provides a 3D center that is projected to the image plane to query local ViT decoder features. The queried tokens are fused with geometry-aware position embeddings and processed by a stack of PTv3 blocks for spatial reasoning. A lightweight Gaussian Head then predicts residual offsets that refine the initial 3DGS template.}
    \label{fig:additional-workflow}
    \vspace{-0.3cm}
\end{figure}

Figure \ref{fig:additional-workflow} presents the detailed workflow of the proposed Gaussian Offset Learning, complementing the description in Section 3.5 of the main paper. Given the densified 3DGS template $\mathcal{G}^{\text{Dense}} = \{ G_i^{\text{Dense}} \}_{i=1}^{N}$ and the decoded ViT feature tensor $\mathbf{t}_{de}$, our Gaussian Offset Learning pipeline refines each Gaussian primitive through a sequence of geometry-appearance fusion operations. For each Gaussian $G_i^{\text{Dense}} = (\boldsymbol{\mu}_i,\, \boldsymbol{\alpha}_i,\, \boldsymbol{r}_i,\, \boldsymbol{s}_i,\, \boldsymbol{c}_i)$, we first project its 3D center $\boldsymbol{\mu}_i$ onto the image plane. Let $\tilde{\boldsymbol{\mu}}_i = [\boldsymbol{\mu}_i^\top,\, 1]^\top \in \mathbb{R}^4$ denote the homogeneous center, and let the camera extrinsic matrix be $\mathbf{P} = [\,\mathbf{R}\mid \mathbf{t}\,] \in \mathbb{R}^{3\times 4}$ with rotation $\mathbf{R}$ and translation $\mathbf{t}$, and intrinsic matrix $\mathbf{K} \in \mathbb{R}^{3 \times 3}$. The homogeneous image coordinate $\tilde{\boldsymbol{p}}_i \in \mathbb{R}^3$ is obtained by
\begin{equation}
\tilde{\boldsymbol{p}}_i
=
\mathbf{K}\mathbf{P}\,\tilde{\boldsymbol{\mu}}_i
=
\begin{bmatrix}
\tilde{u}_i \\
\tilde{v}_i \\
\tilde{w}_i
\end{bmatrix},
\end{equation}
where $\tilde{u}_i$, $\tilde{v}_i$, and $\tilde{w}_i$ denote the homogeneous pixel coordinates. The final 2D pixel position $\boldsymbol{p}_i = (u_i, v_i)^\top$ on the image plane is obtained by inhomogeneous normalization:
\begin{equation}
    u_i = \frac{\tilde{u}_i}{\tilde{w}_i}, 
    \qquad
    v_i = \frac{\tilde{v}_i}{\tilde{w}_i}.
\end{equation}

These 3D centers are also fed into a position embedding network to generate the corresponding \emph{Gaussian position tokens}, providing geometry-aware descriptors for each primitive. 
In parallel, the feature map $\mathbf{t}_{de} \in \mathbb{R}^{4 \times 4 \times 768}$ is reshaped into a grid of local descriptors,  from which we extract the feature $\mathbf{F}_i$ corresponding to $\boldsymbol{p}_i$. This queried feature serves as the \emph{queried token} shown in the diagram.
The Gaussian position token and queried image token are then fused and passed through a stack of $M$ PointTransformerV3 (PTv3) blocks, which model geometric relations, neighborhood context, and long-range interactions among Gaussians. This produces an enhanced latent representation for each primitive. Finally, the encoded features are fed into a lightweight Gaussian Head, implemented as a small MLP, which predicts the residual parameter offsets $\Delta G_i = (\Delta\boldsymbol{\mu}_i,\,  \Delta\boldsymbol{\alpha}_i,\,  \Delta\boldsymbol{r}_i,\,  \Delta\boldsymbol{s}_i,\,  \Delta\boldsymbol{c}_i)$.

\section{Additional Zero-Shot Visualizations on DTU}
\begin{figure}[t]
    \centering
    \vspace{-0.1cm}
    \includegraphics[width=1.0\linewidth]{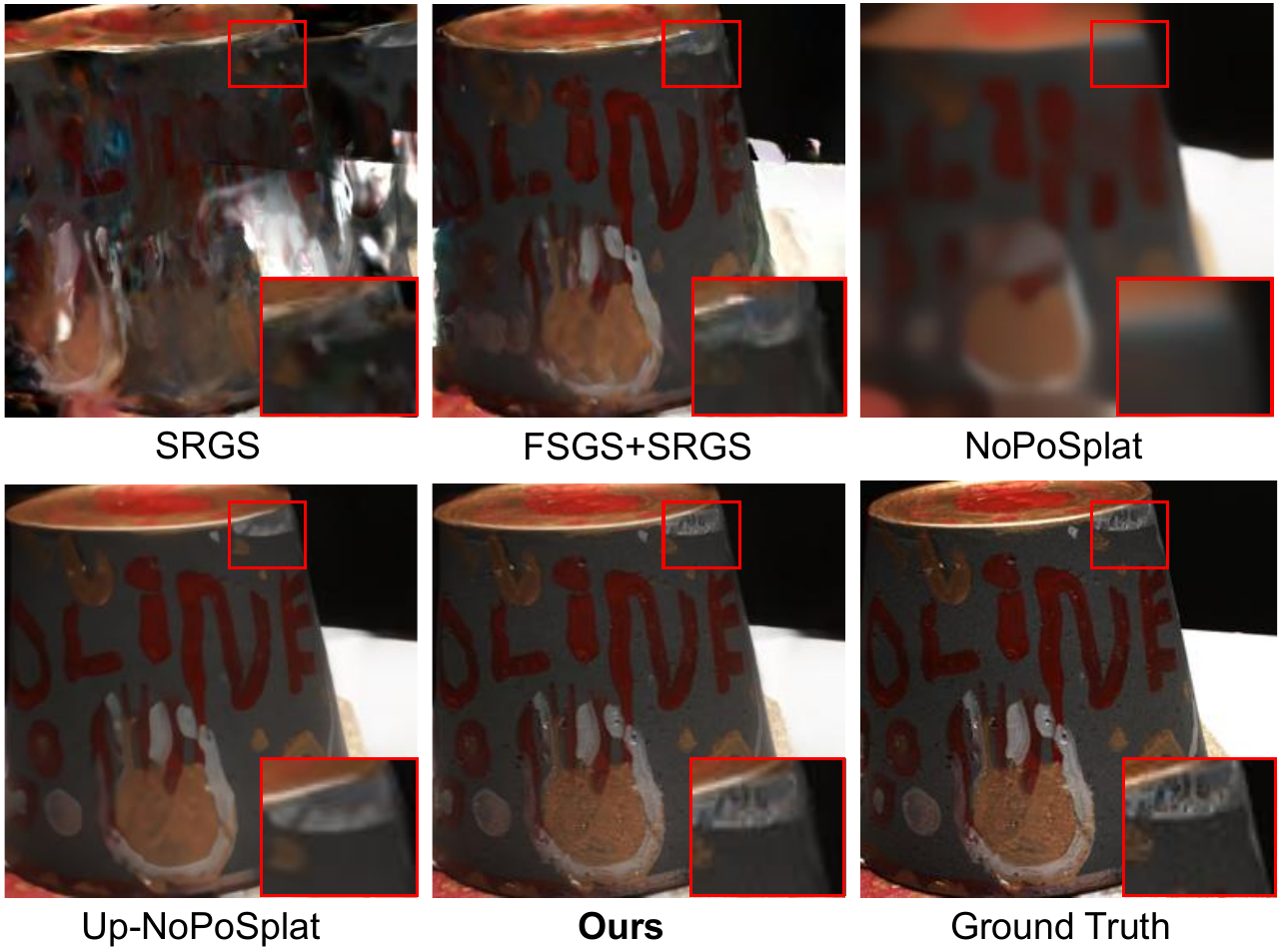}
    \vspace{-0.3cm}
    \caption{\textbf{Zero-shot qualitative comparison on the DTU dataset.} Per-scene optimization and feed-forward baselines show blurring and geometric artifacts, while SR3R recovers significantly sharper textures and consistent geometry, highlighting its strong generalization to unseen scenes.}
    \label{fig:additional-3}
    \vspace{-0.1cm}
\end{figure}

The main paper reports quantitative zero-shot results on the DTU dataset, demonstrating that SR3R achieves the highest accuracy among both feed-forward and per-scene optimization methods. To complement these quantitative findings, Figure \ref{fig:additional-3} presents additional qualitative comparisons on DTU. As can be seen, both feed-forward and optimization-based baselines struggle under sparse LR inputs. SRGS and FSGS+SRGS exhibit strong geometric distortions and severe texture degradation, while NoPoSplat and its upsampled variant produce blurry or unstable high-frequency details. In contrast, SR3R reconstructs sharper textures, clearer boundaries, and substantially more stable geometry, consistent with the improvements observed on other datasets. These visualizations further validate SR3R’s strong cross-scene generalization and its ability to recover fine 3D structure on completely unseen scenes.

\section{Additional Zero-shot Evaluation on ScanNet++}
\label{sec:scannet++}

To further validate the generalization ability of SR3R, we perform an additional zero-shot experiment on the ScanNet++ dataset \cite{yeshwanth2023scannet++}, which contains indoor scenes with different camera motion and scene types from the RE10K. The experimental setup follows the same protocol as in the main paper: all feed-forward models, including SR3R and the baselines, are trained on RE10K and directly tested on ScanNet++ without any fine-tuning. The per-scene optimization methods SRGS and FSGS+SRGS are evaluated using scene-specific optimization.

Table \ref{tab:scannet++} shows that SR3R achieves the highest performance across all metrics, outperforming both feed-forward baselines and per-scene optimization methods. This experiment further demonstrates the strong cross-scene generalization of SR3R and its ability to recover high-frequency geometry and appearance on completely unseen datasets.

\begin{figure}[ht]
    \centering
    \vspace{-0.1cm}
    \includegraphics[width=1.0\linewidth]{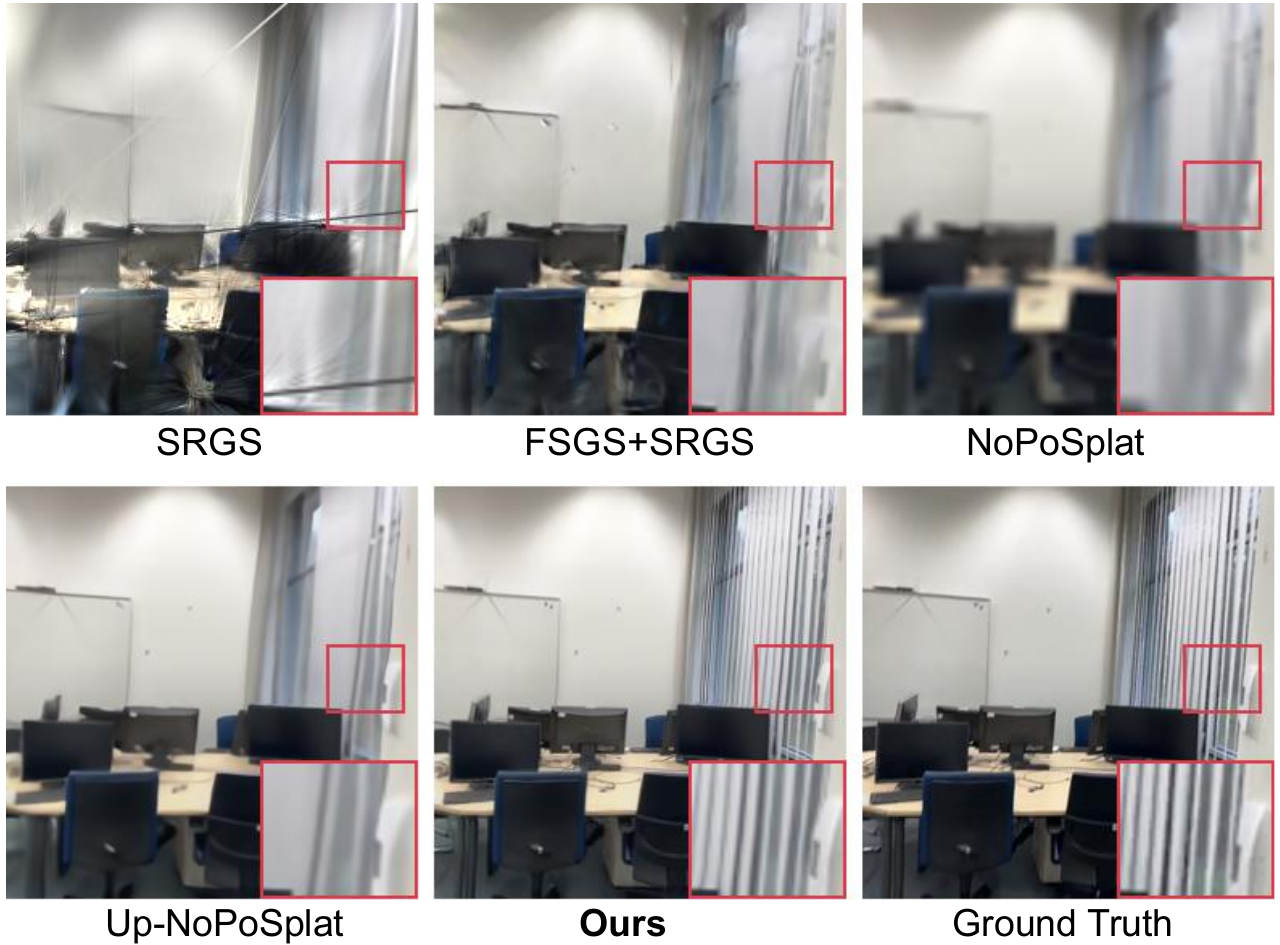}
    \vspace{-0.3cm}
    \caption{\textbf{Zero-shot qualitative comparison on the ScanNet++ dataset.} Per-scene optimization and feed-forward baselines show blurring and geometric artifacts, while SR3R recovers significantly sharper textures and consistent geometry, highlighting its strong generalization to unseen scenes.}
    \label{fig:additional-4}
    \vspace{-0.1cm}
\end{figure}

Figure \ref{fig:additional-4} presents the qualitative comparisons on ScanNet++. As shown, the per-scene optimization methods SRGS and FSGS+SRGS exhibit strong geometric distortions and unstable shading artifacts under sparse LR inputs. Feed-forward baselines, including NoPoSplat and its upsampled variant, remain overly smooth and fail to recover high-frequency textures such as fine surface patterns or sharp edges. In contrast, SR3R reconstructs clearer textures, cleaner boundaries, and more stable geometry, closely matching the ground-truth appearance. These results further validate the strong cross-dataset generalization of SR3R.

\begin{table}[t]
    \renewcommand{\arraystretch}{1.2}
    \footnotesize
    \centering
    \caption{\textbf{Zero-shot generalization results from RE10K to Scanet++.} Feed-forward models are trained on RE10K and tested on Scanet++ without fine-tuning. SRGS and FSGS+SRGS use per-scene optimization. SR3R delivers the best reconstruction quality while remaining significantly faster than optimization-based methods. \textbf{Bold} indicates the best results.} %and \underline{underline} the second best.}
    \label{tab:scannet++}
    \vspace{-0.1cm}
    \resizebox{\linewidth}{!}
    {
    \begin{tabular}{l cccc}
    \Xhline{1pt}
    \multirow{2}{*}{\textbf{Method}} & \multicolumn{4}{c}{\textbf{RE10K $\rightarrow$ Scanet++}}  \\
    \cline{2-5} 
          & PSNR $\uparrow$   & SSIM $\uparrow$  & LPIPS $\downarrow$ & Rec. Time $\downarrow$  \\ 
    \Xhline{1pt}
    SRGS \cite{feng2024srgs}   &12.542 &0.455 &0.502 & 240s\\
    FSGS+SRGS \cite{zhu2024fsgs} &16.514 &0.596 &0.409  &280s\\
    \hline
    NopoSplat \cite{Ye2025}  &18.284 &0.578 &0.421 &0.01s \\ %0.009
    Up-Noposplat &20.870 &0.696 &0.303 &{0.16s}\\ %0.013
    %\hline
    \textbf{Ours (NopoSplat) } &\textbf{21.743} &\textbf{0.739} &\textbf{0.256} & 1.69s\\ %1.699
    \Xhline{1pt}
    \end{tabular}
    }
\vspace{-0.2cm}
\end{table}

%\section{Additional Zero-Shot Visualizations on ScanNet++}

\section{Additional Qualitative Comparisons}
\begin{figure*}[t]
    \centering
    \includegraphics[width=1.0\linewidth]{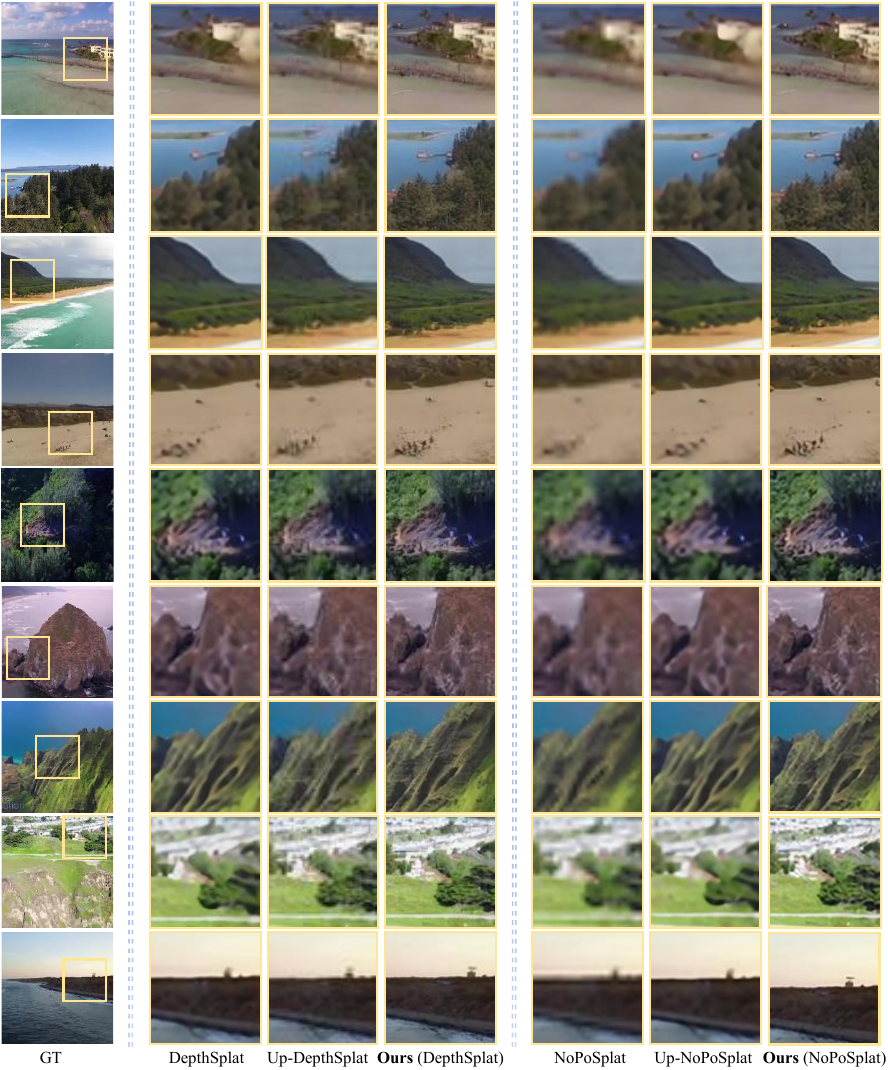}
    \caption{\textbf{Qualitative comparison with SOTA feed-forward 3DGS reconstruction methods on the ACID dataset.} SR3R delivers significantly sharper details and more stable geometry than DepthSplat, NoPoSplat, and their upsampled variants, consistently improving reconstruction quality across different 3DGS backbones under sparse LR inputs.}
    \label{fig:additional-1}
\end{figure*}

\begin{figure*}[t]
    \centering
    \includegraphics[width=1.0\linewidth]{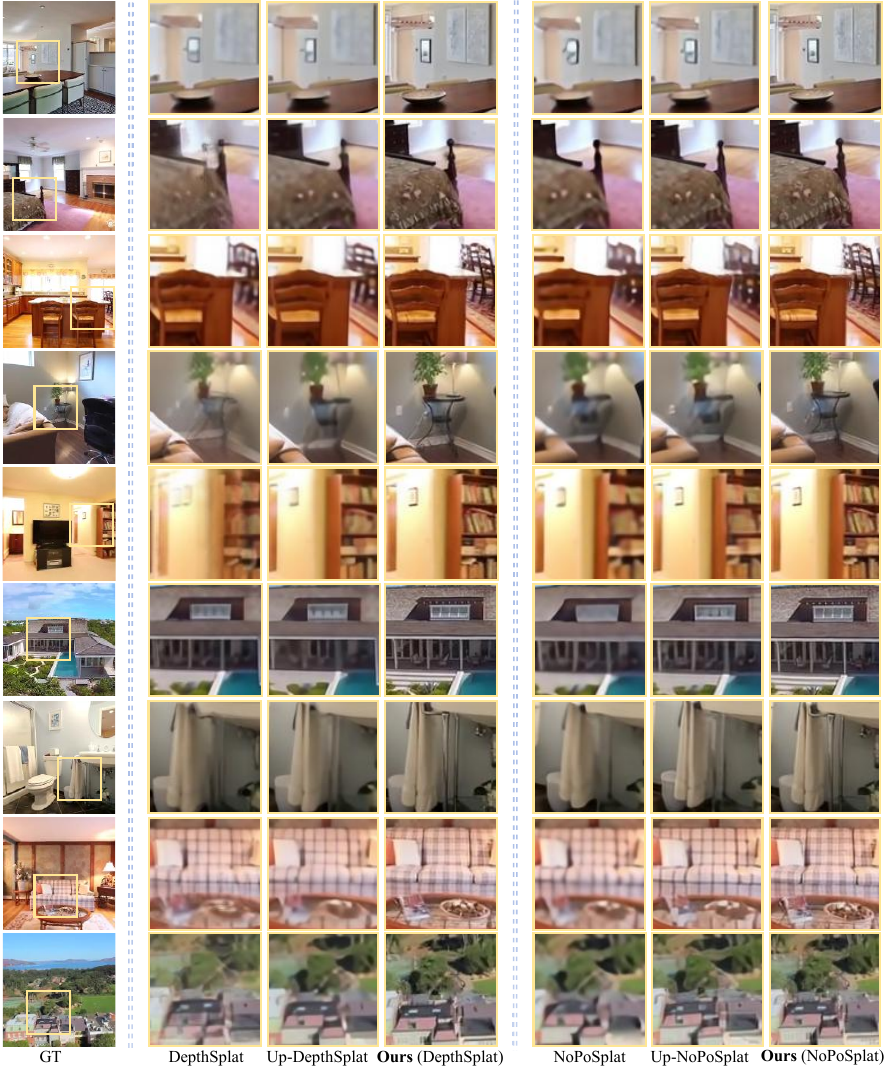}
    \caption{\textbf{Qualitative comparison with SOTA feed-forward 3DGS reconstruction methods on the RE10k dataset.} SR3R delivers significantly sharper details and more stable geometry than DepthSplat, NoPoSplat, and their upsampled variants, consistently improving reconstruction quality across different 3DGS backbones under sparse LR inputs.}
    \label{fig:additional-2}
\end{figure*}

To complement the qualitative comparisons in Figure 3 of the main paper, we provide additional visual results in Figures \ref{fig:additional-1} and \ref{fig:additional-2}. These examples follow the same evaluation protocol and compare SR3R with NoPoSplat, DepthSplat, and their upsampled-input variants. Across a wide range of scenes, the same patterns observed in the main paper consistently hold: feed-forward baselines exhibit noticeable blurring, texture flattening, and geometric instability, while their upsampled variants still fail to recover reliable high-frequency structure. In contrast, our SR3R produces sharper textures, clearer boundaries, and more stable geometry across views. The improvements are consistent for both backbones, demonstrating that our offset-based refinement and cross-view fusion robustly enhance 3D-specific high-frequency reconstruction under sparse LR inputs. These extended visualizations further substantiate the conclusions drawn in the main paper and highlight the reliability of SR3R across diverse scenes.

% {
%     \small
%     \bibliographystyle{ieeenat_fullname}
%     \bibliography{main}
% }


{
    \small
    \bibliographystyle{ieeenat_fullname}
    \bibliography{main}

@inproceedings{Ye2025,
  title={No Pose, No Problem: Surprisingly Simple 3D Gaussian Splats from Sparse Unposed Images},
  author={Ye, Botao and Liu, Sifei and Xu, Haofei and Li, Xueting and others},
  booktitle={The Thirteenth International Conference on Learning Representations},
  year={2025}
}

@inproceedings{xu2025depthsplat,
  title={Depthsplat: Connecting gaussian splatting and depth},
  author={Xu, Haofei and Peng, Songyou and Wang, Fangjinhua and Blum, Hermann and Barath, Daniel and Geiger, Andreas and Pollefeys, Marc},
  booktitle={Proceedings of the Computer Vision and Pattern Recognition Conference},
  pages={16453--16463},
  year={2025}
}

@inproceedings{wan2025s2gaussian,
  title={S2Gaussian: Sparse-View Super-Resolution 3D Gaussian Splatting},
  author={Wan, Yecong and Shao, Mingwen and Cheng, Yuanshuo and Zuo, Wangmeng},
  booktitle={Proceedings of the Computer Vision and Pattern Recognition Conference},
  pages={711--721},
  year={2025}
}

@inproceedings{zhu2024fsgs,
  title={Fsgs: Real-time few-shot view synthesis using gaussian splatting},
  author={Zhu, Zehao and Fan, Zhiwen and Jiang, Yifan and Wang, Zhangyang},
  booktitle={European Conference on Computer Vision},
  pages={145--163},
  year={2024}
}

@String(CVPR= {IEEE Conf. Comput. Vis. Pattern Recog.})

@String(ECCV= {Eur. Conf. Comput. Vis.})

@String(TOG= {ACM Trans. Graph.})

@String(CVPR  = {CVPR})

@String(ECCV  = {ECCV})

@String(TOG   = {ACM TOG})

@Article{3Dgaussians,
      author       = {Kerbl, Bernhard and Kopanas, Georgios and Leimk{\"u}hler, Thomas and Drettakis, George},
      title        = {3D Gaussian Splatting for Real-Time Radiance Field Rendering},
      journal      = {ACM Transactions on Graphics},
      number       = {4},
      volume       = {42},
      month        = {July},
      year         = {2023}
}

@inproceedings{CROC,
    author    = {Yoon, Youngho and Yoon, Kuk-Jin},
    title     = {Cross-Guided Optimization of Radiance Fields With Multi-View Image Super-Resolution for High-Resolution Novel View Synthesis},
    booktitle = {Proceedings of the IEEE/CVF Conference on Computer Vision and Pattern Recognition (CVPR)},
    month     = {June},
    year      = {2023},
    pages     = {12428-12438}
}

@InProceedings{MipSplatting,
    author    = {Yu, Zehao and Chen, Anpei and Huang, Binbin and Sattler, Torsten and Geiger, Andreas},
    title     = {Mip-Splatting: Alias-free 3D Gaussian Splatting},
    booktitle = {Proceedings of the IEEE/CVF Conference on Computer Vision and Pattern Recognition (CVPR)},
    month     = {June},
    year      = {2024},
    pages     = {19447-19456}
}

@INPROCEEDINGS{Swinir,
  author={Liang, Jingyun and Cao, Jiezhang and Sun, Guolei and Zhang, Kai and Van Gool, Luc and Timofte, Radu},
  booktitle={IEEE/CVF International Conference on Computer Vision Workshops}, 
  title={SwinIR: Image Restoration Using Swin Transformer},
  year={2021},
  pages={1833-1844}
}

@INPROCEEDINGS{SSIM,
  author={Wang, Z. and Simoncelli, E.P. and Bovik, A.C.},
  booktitle={The Thrity-Seventh Asilomar Conference on Signals, Systems \& Computers}, 
  title={Multiscale structural similarity for image quality assessment},
  year={2003},
  volume={2},
  pages={1398-1402 Vol.2}
}

@article{gaussiansr,
  title={GaussianSR: 3D Gaussian Super-Resolution with 2D Diffusion Priors},
  author={Yu, Xiqian and Zhu, Hanxin and He, Tianyu and Chen, Zhibo},
  journal={arXiv preprint arXiv:2406.10111},
  year={2024}
}

@InProceedings{EDSR,
  author = {Lim, Bee and Son, Sanghyun and Kim, Heewon and Nah, Seungjun and Lee, Kyoung Mu},
  title = {Enhanced Deep Residual Networks for Single Image Super-Resolution},
  booktitle = {Proceedings of the IEEE/CVF Conference on Computer Vision and Pattern Recognition (CVPR)},
  year = {2017}
}

@inproceedings{rcan,
    title={Image Super-Resolution Using Very Deep Residual Channel Attention Networks},
    author={Zhang, Yulun and Li, Kunpeng and Li, Kai and Wang, Lichen and Zhong, Bineng and Fu, Yun},
    booktitle={European Conference on Computer Vision (ECCV)},
    year={2018}
}

@article{SR3,
title={Image super-resolution via iterative refinement},
author={Saharia, Chitwan and Ho, Jonathan and Chan, William and Salimans, Tim and Fleet, David J and Norouzi, Mohammad},
journal={arXiv:2104.07636},
year={2021}
}

@InProceedings{ESRGAN,
    author = {Wang, Xintao and Yu, Ke and Wu, Shixiang and Gu, Jinjin and Liu, Yihao and Dong, Chao and Qiao, Yu and Loy, Chen Change},
    title = {ESRGAN: Enhanced super-resolution generative adversarial networks},
    booktitle = {European Conference on Computer Vision (ECCV)},
    month = {September},
    year = {2018}
}

@article{resshift,
  title={Resshift: Efficient diffusion model for image super-resolution by residual shifting},
  author={Yue, Zongsheng and Wang, Jianyi and Loy, Chen Change},
  journal={Advances in Neural Information Processing Systems},
  volume={36},
  year={2024}
}

@INPROCEEDINGS{lpip,
  author={Zhang, Richard and Isola, Phillip and Efros, Alexei A. and Shechtman, Eli and Wang, Oliver},
  booktitle={Proceedings of the IEEE/CVF Conference on Computer Vision and Pattern Recognition (CVPR)}, 
  title={The Unreasonable Effectiveness of Deep Features as a Perceptual Metric}, 
  year={2018},
  pages={586-595}
}

@misc{FlashVSR,
      title={FlashVSR: Towards Real-Time Diffusion-Based Streaming Video Super-Resolution}, 
      author={Junhao Zhuang and Shi Guo and Xin Cai and Xiaohui Li and Yihao Liu and Chun Yuan and Tianfan Xue},
      year={2025},
      eprint={2510.12747},
      archivePrefix={arXiv},
      primaryClass={cs.CV},
      url={https://arxiv.org/abs/2510.12747}, 
}

@article{xu2024videogigagan,
      title={VideoGigaGAN: Towards Detail-rich Video Super-Resolution}, 
      author={Yiran Xu and Taesung Park and Richard Zhang and Yang Zhou and Eli Shechtman and Feng Liu and Jia-Bin Huang and Difan Liu},
      year={2024},
      eprint={2404.12388},
      archivePrefix={arXiv},
      primaryClass={cs.CV}
  }

@inproceedings{SRGAN,
  title={Photo-realistic single image super-resolution using a generative adversarial network},
  author={Ledig, Christian and Theis, Lucas and Husz{\'a}r, Ferenc and Caballero, Jose and Cunningham, Andrew and Acosta, Alejandro and Aitken, Andrew and Tejani, Alykhan and Totz, Johannes and Wang, Zehan and others},
  booktitle={Proceedings of the IEEE/CVF Conference on Computer Vision and Pattern Recognition (CVPR)},
  pages={4681--4690},
  year={2017}
}

@inproceedings{idm,
  title={Implicit diffusion models for continuous super-resolution},
  author={Gao, Sicheng and Liu, Xuhui and Zeng, Bohan and Xu, Sheng and Li, Yanjing and Luo, Xiaoyan and Liu, Jianzhuang and Zhen, Xiantong and Zhang, Baochang},
  booktitle={Proceedings of the IEEE/CVF Conference on Computer Vision and Pattern Recognition (CVPR)},
  pages={10021--10030},
  year={2023}
}

@InProceedings{FSRCNN,
author="Dong, Chao
and Loy, Chen Change
and Tang, Xiaoou",
editor="Leibe, Bastian
and Matas, Jiri
and Sebe, Nicu
and Welling, Max",
title="Accelerating the Super-Resolution Convolutional Neural Network",
booktitle="European Conference on Computer Vision (ECCV)",
year="2016",
publisher="Springer International Publishing",
address="Cham",
pages="391--407",
isbn="978-3-319-46475-6"
}

@article{hat,
  title={HAT: Hybrid Attention Transformer for Image Restoration},
  author={Chen, Xiangyu and Wang, Xintao and Zhang, Wenlong and Kong, Xiangtao and Qiao, Yu and Zhou, Jiantao and Dong, Chao},
  journal={arXiv preprint arXiv:2309.05239},
  year={2023}
}

@article{shi2025mmgs,
  title={MMGS: Multi-model synergistic Gaussian splatting for sparse view synthesis},
  author={Shi, Changyue and Yang, Chuxiao and Hu, Xinyuan and Yang, Yan and Ding, Jiajun and Tan, Min},
  journal={Image and Vision Computing},
  pages={105512},
  year={2025},
  publisher={Elsevier}
}

@article{EDT,
  title={On Efficient Transformer and Image Pre-training for Low-level Vision},
  author={Li, Wenbo and Lu, Xin and Qian, Shengju and Lu, Jiangbo and Zhang, Xiangyu and Jia, Jiaya},
  journal={arXiv preprint arXiv:2112.10175},
  year={2021}
}

@inproceedings{SuperGaussian,
  title = {SuperGaussian: Repurposing Video Models for 3D Super Resolution},
  author = {Shen, Yuan and Ceylan, Duygu and Guerrero, Paul and Xu, Zexiang and Mitra, {Niloy J.} and Wang, Shenlong and Fr{"u}hst{"u}ck, Anna},
  booktitle = {European Conference on Computer Vision (ECCV)},
  year = {2024}
}

@article{feng2024srgs,
  title={Srgs: Super-resolution 3d gaussian splatting},
  author={Feng, Xiang and He, Yongbo and Wang, Yubo and Yang, Yan and Li, Wen and Chen, Yifei and others},
  journal={arXiv preprint arXiv:2404.10318},
  year={2024}
}

@inproceedings{basicvsr,
  title={Basicvsr: The search for essential components in video super-resolution and beyond},
  author={Chan, Kelvin CK and Wang, Xintao and Yu, Ke and Dong, Chao and Loy, Chen Change},
  booktitle={Proceedings of the IEEE/CVF Conference on Computer Vision and Pattern Recognition (CVPR)},
  pages={4947--4956},
  year={2021}
}

@article{supergs,
  title={SuperGS: Super-Resolution 3D Gaussian Splatting via Latent Feature Field and Gradient-guided Splitting},
  author={Xie, Shiyun and Wang, Zhiru and Zhu, Yinghao and Pan, Chengwei},
  journal={arXiv preprint arXiv:2410.02571},
  year={2024}
}

@article{sequence,
  title={Sequence Matters: Harnessing Video Models in 3D Super-Resolution},
  author={Ko, Hyun-kyu and Park, Dongheok and Park, Youngin and Lee, Byeonghyeon and Han, Juhee and Park, Eunbyung},
  journal={arXiv preprint arXiv:2412.11525},
  year={2024}
}

@inproceedings{corgs,
  title={CoR-GS: sparse-view 3D Gaussian splatting via co-regularization},
  author={Zhang, Jiawei and Li, Jiahe and Yu, Xiaohan and Huang, Lei and Gu, Lin and Zheng, Jin and Bai, Xiao},
  booktitle={European Conference on Computer Vision (ECCV)},
  pages={335--352},
  year={2024},
  organization={Springer}
}

@inproceedings{gaussianpro,
  title={Gaussianpro: 3d gaussian splatting with progressive propagation},
  author={Cheng, Kai and Long, Xiaoxiao and Yang, Kaizhi and Yao, Yao and Yin, Wei and Ma, Yuexin and Wang, Wenping and Chen, Xuejin},
  booktitle={Forty-first International Conference on Machine Learning},
  year={2024}
}

@inproceedings{depthsplat,
  title={Depthsplat: Connecting gaussian splatting and depth},
  author={Xu, Haofei and Peng, Songyou and Wang, Fangjinhua and Blum, Hermann and Barath, Daniel and Geiger, Andreas and Pollefeys, Marc},
  booktitle={Proceedings of the Computer Vision and Pattern Recognition Conference},
  pages={16453--16463},
  year={2025}
}

@inproceedings{pixelsplat,
  title={pixelsplat: 3d gaussian splats from image pairs for scalable generalizable 3d reconstruction},
  author={Charatan, David and Li, Sizhe Lester and Tagliasacchi, Andrea and Sitzmann, Vincent},
  booktitle={Proceedings of the IEEE/CVF conference on computer vision and pattern recognition},
  pages={19457--19467},
  year={2024}
}

@inproceedings{chen2024mvsplat,
  title={Mvsplat: Efficient 3d gaussian splatting from sparse multi-view images},
  author={Chen, Yuedong and Xu, Haofei and Zheng, Chuanxia and Zhuang, Bohan and Pollefeys, Marc and Geiger, Andreas and Cham, Tat-Jen and Cai, Jianfei},
  booktitle={European Conference on Computer Vision},
  pages={370--386},
  year={2024},
  organization={Springer}
}

@article{super-nerf,
  title={Super-nerf: View-consistent detail generation for nerf super-resolution},
  author={Han, Yuqi and Yu, Tao and Yu, Xiaohang and Xu, Di and Zheng, Binge and Dai, Zonghong and Yang, Changpeng and Wang, Yuwang and Dai, Qionghai},
  journal={IEEE Transactions on Visualization and Computer Graphics},
  year={2024},
  publisher={IEEE}
}

@inproceedings{analytic,
  title={Analytic-splatting: Anti-aliased 3d gaussian splatting via analytic integration},
  author={Liang, Zhihao and Zhang, Qi and Hu, Wenbo and Zhu, Lei and Feng, Ying and Jia, Kui},
  booktitle={European Conference on Computer Vision (ECCV)},
  pages={281--297},
  year={2024},
  organization={Springer}
}

@article{nopo,
  title={No pose, no problem: Surprisingly simple 3d gaussian splats from sparse unposed images},
  author={Ye, Botao and Liu, Sifei and Xu, Haofei and Li, Xueting and others},
  journal={arXiv preprint arXiv:2410.24207},
  year={2024}
}

@article{weng2025gaussianlens,
  title={GaussianLens: Localized High-Resolution Reconstruction via On-Demand Gaussian Densification},
  author={Weng, Yijia and Wang, Zhicheng and Peng, Songyou and Xie, Saining and Zhou, Howard and Guibas, Leonidas J},
  journal={arXiv preprint arXiv:2509.25603},
  year={2025}
}

@article{wang2025styl3r,
  title={Styl3R: Instant 3D Stylized Reconstruction for Arbitrary Scenes and Styles},
  author={Wang, Peng and Liu, Xiang and Liu, Peidong},
  journal={arXiv preprint arXiv:2505.21060},
  year={2025}
}

@article{liu2025stylos,
  title={Stylos: Multi-View 3D Stylization with Single-Forward Gaussian Splatting},
  author={Liu, Hanzhou and Huang, Jia and Lu, Mi and Saripalli, Srikanth and Jiang, Peng},
  journal={arXiv preprint arXiv:2509.26455},
  year={2025}
}

@article{xu2025siu3r,
  title={SIU3R: Simultaneous Scene Understanding and 3D Reconstruction Beyond Feature Alignment},
  author={Xu, Qi and Wei, Dongxu and Zhao, Lingzhe and Li, Wenpu and Huang, Zhangchi and Ji, Shunping and Liu, Peidong},
  journal={arXiv preprint arXiv:2507.02705},
  year={2025}
}

@inproceedings{DTU,
  title={Large scale multi-view stereopsis evaluation},
  author={Jensen, Rasmus and Dahl, Anders and Vogiatzis, George and Tola, Engin and Aan{\ae}s, Henrik},
  booktitle={Proceedings of the IEEE conference on computer vision and pattern recognition},
  pages={406--413},
  year={2014}
}

@inproceedings{yeshwanth2023scannet++,
  title={Scannet++: A high-fidelity dataset of 3d indoor scenes},
  author={Yeshwanth, Chandan and Liu, Yueh-Cheng and Nie{\ss}ner, Matthias and Dai, Angela},
  booktitle={Proceedings of the IEEE/CVF International Conference on Computer Vision},
  pages={12--22},
  year={2023}
}

@article{Re10k,
  title={Stereo magnification: learning view synthesis using multiplane images},
  author={Zhou, Tinghui and Tucker, Richard and Flynn, John and Fyffe, Graham and Snavely, Noah},
  journal={ACM Transactions on Graphics (TOG)},
  volume={37},
  number={4},
  pages={1--12},
  year={2018},
  publisher={ACM New York, NY, USA}
}

@inproceedings{ACID,
  title={Infinite nature: Perpetual view generation of natural scenes from a single image},
  author={Liu, Andrew and Tucker, Richard and Jampani, Varun and Makadia, Ameesh and Snavely, Noah and Kanazawa, Angjoo},
  booktitle={Proceedings of the IEEE/CVF International Conference on Computer Vision},
  pages={14458--14467},
  year={2021}
}
}

\end{document}